\documentclass[final,3p,times,twocolumn]{elsarticle}
\usepackage{etoolbox}
\newbool{submit}
\booltrue{submit}

\usepackage{placeins}
\usepackage{graphicx}
\usepackage{ifpdf}
\ifpdf
   \usepackage{epstopdf}
\else
   \usepackage{graphicx}
\fi

\usepackage[acronym]{glossaries}
\usepackage[table,usenames,dvipsnames]{xcolor}
\usepackage{nameref}
\usepackage{tabularx}

\usepackage{amssymb}

\usepackage{mathptmx} \usepackage{amssymb,amsmath,amstext}
\usepackage[T1]{fontenc}
\usepackage{textcomp}
\DeclareMathAlphabet{\bm}{OT1}{ptm}{b}{it}

\usepackage[format=hang,font=small,labelfont=bf]{caption}
\usepackage{subscript}
\usepackage{multirow}
\usepackage{algorithm}

\usepackage{algorithmic}
\usepackage{listings}

\usepackage{subfigure}

\usepackage{relsize}
\usepackage{lscape}
\usepackage[abs]{overpic}

\ifbool{submit}
{
  
}
{
  \usepackage[colorinlistoftodos]{todonotes}
  \usepackage[nolists]{endfloat}
  
  \usepackage[marginparwidth=1.2in]{geometry}
}

\usepackage[hyphens]{url}
\usepackage[hidelinks]{hyperref}

\usepackage{natbib}
\biboptions{sort&compress}
\definecolor{lightgray}{gray}{0.95}

\newcolumntype{Z}{>{\centering\arraybackslash}X}
\newcolumntype{L}{>{\raggedright\arraybackslash}X}

\DeclareMathAlphabet{\mathcal}{OMS}{cmsy}{m}{n}
\newcommand{\degree}{\ensuremath{^{\circ}}}

\journal{Comput. Meth. Prog. Bio.}

\makeatletter
\let\x@caption\caption \def\x@@caption[#1]#2{\x@caption[{#1}]{\textbf{#1}. #2}} \def\x@@@caption#1{\x@caption[{#1}]{#1}} \def\caption{\@ifnextchar[\x@@caption\x@@@caption} \makeatother

\renewcommand{\vec}[1]{\ensuremath{\mathbf{#1}}}

\newcommand{\vy}{\vec{y}}
\renewcommand*{\UrlFont}{\ttfamily\smaller\relax}
\newcommand*{\doi}{\begingroup\catcode`_=12 \addurldoi}
\newcommand{\addurldoi}[1]{doi: \href{http://dx.doi.org/#1}{\UrlFont#1}\endgroup}

\newcommand{\bitem}[1]{\item \textbf{#1}.}

\makeatletter{}\newacronym[first={\emph{MBIS} (Multivariate Bayesian Image Segmentation tool)}]{mbis}{MBIS}{Multivariate Bayesian Image Segmentation tool}

\newacronym[first={\emph{FLIRT} (FMRIB's Linear Image Registration Tool \cite{jenkinson_improved_2002})}]{flirt}{FLIRT}{FMRIB's Linear Image Registration Tool}

\newacronym[first={\emph{SPM} (Statistical Parametric Mapping, The Wellcome Dept.
  of Imaging Neuroscience, London, UK \cite{ashburner_unified_2005})}]{spm}{SPM}{Statistical Parametric Mapping}

\newacronym[first={\emph{ANTS} (Advanced Normalization Tools \cite{avants_ants:_2013})}]{ants}{ANTS}{Advanced Normalization tools}

\newacronym[first={\emph{ITK} (the Insight Registration \& Segmentation Toolkit,
  \url{http://www.itk.org})}]{itk}{ITK}{Insight Registration \& Segmentation Toolkit}

\newacronym[first={\emph{FAST} (Fast Automated Segmentation Tool \cite{zhang_segmentation_2001})}]{fast}{FAST}{Fast Automated Segmentation Tool}

\newacronym[first={\emph{BET} (Brain Extraction Tool \cite{smith_fast_2002})}]{bet}{BET}{Brain Extraction Tool}

\newacronym[first={\emph{EMS} (Expectation-Maximization Segmentation \cite{van_leemput_unifying_2003})}]{leemput_ems}{EMS}{Expectation-Maximization Segmentation}

\newglossaryentry{nipype}{  name        = {\emph{nipype}},  description = {Neuroimaging in Python, Pipelines and Interfaces},  first       = {\emph{nipype} (Neuroimaging in Python, Pipelines, and Interfaces \cite{gorgolewski_nipype:_2011})},}

\newacronym[first={\emph{N4ITK} \cite{tustison_n4itk:_2010}}]{n4itk}{N4ITK}{N4ITK}

\newacronym{miccai}{MICCAI}{Medical Image Computing and Computer Assisted Intervention}
\newacronym{multivariate_gaussian}{MMG}{mixture of multivariate Gaussian distributions}
\newacronym{tpm}{TPM}{tissue probability map}
\newacronym{markov}{MRF}{hidden Markov random field}
\newacronym{graph_cuts}{GC}{graph-cuts}
\newacronym{icm}{ICM}{iterative conditional models}
\newacronym{icv}{ICV}{intra-cranial volume}
\newacronym{e_m}{EM}{expectation-maximization}
\newacronym{map}{MAP}{maximum a posteriori}
\newacronym{mri}{MRI}{magnetic resonance imaging}
\newacronym{t1}{T1w}{T1-weighted}
\newacronym{t2}{T2w}{T2-weighted}
\newacronym{pd}{PDw}{proton-density-weighted}
\newacronym{mt}{MT}{magnetization transfer imaging}
\newacronym{wm}{WM}{white matter}
\newacronym{gm}{GM}{gray matter}
\newacronym{csf}{CSF}{cerebrospinal fluid}
\newacronym{pv}{PV}{partial volume}
\newacronym{fsl}{FSL}{FMRIB Software Library 4.1}
\newacronym{gpl}{GPL}{GNU General Public License}
\newacronym{montecarlo}{MC}{Monte-Carlo}
\newacronym{lm}{LM}{Levenberg-Marquardt}
\newacronym{dct}{DCT}{discrete cosine transform}

\newacronym[first={fuzzy similarity index (\emph{fSI})}]{ev_fsi}{fSI}{fuzzy similarity index}
\newacronym[first={similarity index (\emph{SI}})]{ev_si}{SI}{similarity index}
\newacronym[first={true-positive fraction (\emph{TPF})}]{ev_tpf}{TPF}{true-positive fraction}
\newacronym[first={extra fraction (\emph{EF}})]{ev_ef}{EF}{extra fraction}
\newacronym[first={overlap conformity measurement (\emph{OC})}]{ev_oc}{OC}{overlap conformity measurement}
\newacronym[first={Jaccard's index (\emph{JI})}]{ev_ji}{JI}{Jaccard's index}

\makeglossaries

\begin{document}

\begin{frontmatter}

\title{MBIS: Multivariate Bayesian Image Segmentation Tool}

\author[bit,lts5,ciber]{Oscar Esteban\corref{ref1}}
\ead{code@oscaresteban.es}

\author[bit,ciber]{Gert Wollny}
\author[lts5]{Subrahmanyam Gorthi}
\author[bit,ciber]{Mar\'ia-Jes\'us Ledesma-Carbayo}
\author[lts5,chuv]{Jean-Philippe Thiran}
\author[bit,ciber]{Andr\'es Santos}
\author[chuv,lts5]{Meritxell Bach-Cuadra}

\cortext[ref1]{Corresponding author. Tel.: +34 915 495 700 ext.4234 }

\address[bit]{Biomedical Image Technologies (BIT), ETSI Telecomunicaci\'on, Universidad Polit\'ecnica de Madrid, Spain}
\address[lts5]{Signal Processing Laboratory (LTS5), \'Ecole Polytechnique F\'ed\'erale de Lausanne (EPFL), Switzerland}
\address[chuv]{Department of Radiology, Centre d'Imaginerie Biom\'edicale, University Hospital Center and University of Lausanne, Switzerland}
\address[ciber]{Biomedical Research Networking Center in Bioengineering, Biomaterials and Nanomedicine (CIBER-BBN), Spain}

\begin{abstract}
\label{sec:abstract}
\makeatletter{}We present \gls*{mbis}, a clustering tool based on
  the mixture of multivariate normal distributions model.
\Gls*{mbis} supports multichannel bias field correction
  based on a B-spline model.
A second methodological novelty is the inclusion of
  \acrlong*{graph_cuts} optimization for the stationary
  anisotropic \acrlong*{markov} model.
Along with \gls*{mbis}, we release an evaluation framework that
  contains three different experiments on multi-site data.
We first validate the accuracy of segmentation and the estimated
  bias field for each channel.
\Gls*{mbis} outperforms a widely used segmentation tool in a
  cross-comparison evaluation.
The second experiment demonstrates the robustness of results
  on atlas-free segmentation of two image sets from scan-rescan
  protocols on 21 healthy subjects.
Multivariate segmentation is more replicable than the monospectral
  counterpart on \acrlong*{t1} images.
Finally, we provide a third experiment to illustrate how \gls*{mbis}
  can be used in a large-scale study of tissue volume change with
  increasing age in 584 healthy subjects.
This last result is meaningful as multivariate segmentation performs
  robustly without the need for prior knowledge. 
\end{abstract}


\end{frontmatter}

\glsresetall[\acronymtype] 

\makeatletter{}\section{Introduction}
\label{sec:introduction}
Brain tissue segmentation from \gls*{mri} has been one of
  the most challenging problems in computer vision
  applied to biomedical image analysis \citep{kapur_segmentation_1996}.
It is intended to provide precise delineations of \gls*{wm}, \gls*{gm}
  and \gls*{csf} from acquired data.
Brain tissue segmentation is the standpoint of processing schemes in an
  endless number of research studies concerning brain morphology,
  such as quantitative analyses of tissue volumes 
  \citep{mortamet_effects_2005,abe_sex_2010,taki_correlations_2011},
  studies of cortical thickness \citep{fischl_measuring_2000,jones_three-dimensional_2000,
  macdonald_automated_2000}, and voxel-based morphometry
  \citep{wright_voxel-based_1995,paus_structural_1999,
  good_voxel-based_2001,ge_age-related_2002}.
In a clinical context, numerous studies have demonstrated the
  potential use of brain tissue segmentation.
The spatial location of the above key anatomical structures within
  the brain is a requirement for clinical intervention \citep{kikinis_digital_1996} 
  (e.g. radiotherapy planning, surgical planning, and image-guided intervention).
Early applications addressed global conditions; for example \citep{tanabe_tissue_1997}
  used semiautomated segmentation of \gls*{mri} to assess the decrease in 
  total brain tissue and cortical \gls*{gm}, and ventricle enlargement in Alzheimer's
  Disease patients.
Another study \citep{hazlett_cortical_2006} presented an automated methodology
  to identify abnormal increase of the \gls*{gm} volume in individuals
  with autism.
Focal conditions have also been studied, including extra classes in clustering and
  some other adaptations of methods to pathologies, such as automated tumor delineation
  \citep{prastawa_automatic_2003}, lesion detection and volume analyses
  in multiple sclerosis \citep{collins_automated_2001,
  zijdenbos_automatic_1998,zijdenbos_automatic_2002,van_leemput_automated_2001,
  van_leemput_unifying_2003}, and white matter lesions associated
  with age and several conditions like clinically silent stroke, and higher
  systolic blood pressure \citep{anbeek_probabilistic_2004}.
The accurate and automated segmentation of tumor and
  edema in multivariate brain images is an active field of interest in medical
  image analysis, as illustrated by the Challenge on \emph{Multimodal
  Brain Tumor Segmentation} \citep{menze_multimodal_2014} that has been 
  held in conjunction with the last three sessions of the 
  \gls*{miccai} International Conference.

A survey on brain tissue segmentation techniques is reported elsewhere 
  \citep{liew_current_2006}.
Currently popular methodologies can be grouped into three main
  families.
\emph{Deformable model fitting} approaches
  \citep{suri_leaking_2000,yushkevich_user-guided_2006,roura_marga:_2012,delibasis_novel_2013,dang_validation_2013}
  are designed to evolve a number of initial contours
  towards the intensity steps that occur at tissue
  interfaces.
\emph{Atlas-based methods} \citep{gorthi_active_2011}
  use image registration
  to perform a spatial mapping between the actual data
  and an anatomical reference called an atlas.
The atlas is \emph{prior} knowledge on the morphology
  of data, and it generally comprehends a partition previously 
  extracted by any other means (i.e. manual delineation,
  averaging large populations, etc.).
\emph{Clustering or classification} algorithms
  \citep{van_leemput_automated_1999-1,ahmed_modified_2002,
  vrooman_multi-spectral_2007,ji_generalized_2012}
  search for a pixel-wise partition of the image 
  data into a certain number of categories or clusters
  (i.e. \gls*{wm}, \gls*{gm}, and \gls*{csf}).
The partition can be \emph{hard} when each pixel belongs
  to a single cluster or \emph{fuzzy},
  assigning a probability of membership to
  each category, which yields a so-called \gls*{tpm} 
  per class.
These three families of segmentation strategies have often been
  combined to obtain enhanced results.
For instance, deformable models can be initialized
  using contours already located close to the solution 
  sought using atlases.
In clustering methods, \emph{priors} usually take the
  form of precomputed \glspl*{tpm} derived from the atlas.
These prior probability maps can be used just to initialize
  the model, or be integrated throughout the model fitting
  process \citep{ashburner_unified_2005}, simultaneously
  improving the atlas registration at each iteration.
The use of \emph{priors} presents two particular properties.
On one hand, it generally aids the segmentation
  process providing great stability and robustness.
However, it is also suspected to bias results,
  driving the solution somewhat close to the
  population features that underlie the atlas
  \citep{davatzikos_why_2004}.
One further concern about the use of \emph{priors} is
  posed by the need for a spatial mapping of the atlas
  information to the actual data 
  \citep{bookstein_voxel-based_2001,
  ashburner_why_2001}, typically performed through a
  registration process that may not be trivial or
  flawless \citep{crum_zen_2003}.
The unpredictable morphology found in pathologic brains
  discourages the use of atlases extracted from healthy populations.
Conversely, monospectral and strictly data-driven approaches are
  usually very unreliable for pathologic subjects.
For instance, a previous study \citep{prastawa_automatic_2003} updated a
  standard atlas with an approximation of tumor locations for automated
  clustering-based segmentation.
On the other hand, multivariate approaches with outlier detection
  \citep{van_leemput_automated_2001} have been proposed in the
  case of multiple sclerosis derived lesions.

The tool proposed in this work, named \gls*{mbis}, belongs to the sub-group of
  Bayesian classification methods, which have been successfully
  applied to brain tissue segmentation for the last 20 years
  \citep{van_leemput_automated_1999-1}.
Therefore, we will restrict the scope of this paper to this
  sub-group of clustering methods.
Given the maturity of the field, numerous evaluation studies have been reported
  \citep{cuadra_comparison_2005,de_boer_accuracy_2010,roche_convergence_2011},
  along with further refinements or extensions to the original methodologies
  \citep{zhang_segmentation_2001,van_leemput_unifying_2003,
  ashburner_unified_2005,fischl_whole_2002}.
Existing applications of brain tissue segmentation generally
  use \gls*{mri} as input data as a safe, noninvasive,
  and highly precise modality.
Early applications typically selected \acrfull*{t1}
  MPRAGE sequences, mainly for their particularly appropriate
  contrast between soft tissues, and for their wide availability.
The current clinical setup provides a large number
  of different sequences that can be used to
  characterize each voxel of the brain with a vector of
  intensities from each different \gls*{mri} scheme.
In the last decade, we have witnessed an
  explosion of the number of \gls*{mri} sequences widely available, enabling
  the exploration of new observed features and requiring powerful multivariate
  processing and analysis.
Moreover, the vast amount of multi-site data
  that research and clinical routines produce daily,
  necessitates accurate and robust methods to
  perform fully automated segmentation on heterogeneous 
  (in the sense of multi-centric and/or multi-scanner) data
  reliably.

In this paper, we contribute to the field with \gls*{mbis}, an open-source 
  software suite to perform multivariate segmentation on 
  heterogeneous data.
We also present a comprehensive evaluation framework,
  containing several validation experiments on data
  from three publicly-available resources.
The first experiment demonstrates the accuracy of \gls*{mbis} segmenting
  one synthetic dataset, in comparison to \gls*{fast}, a widely-used tool.
The second experiment demonstrates the repeatability of results,
  reporting the disagreement between segmentations of two
  multivariate images of the same subject.
These images correspond to 21 subjects who underwent a scan-rescan
  session with the same \gls*{mri} protocol acquired twice.
The third experiment proves the suitability of \gls*{mbis} on large-scale
  segmentation studies.
We demonstrate the successful application of \gls*{mbis} on a multi-site
  resource of 584 subjects and observe the aging effects over
  tissue volumes.

The manuscript is structured as follows: In \autoref{sec:methods},
  after introducing the theoretical background, we describe the
  particular features of the method implemented by \gls*{mbis},
  highlighting its methodological novelties.
In \autoref{sec:software}, we review the existing software that can be
  used to perform brain tissue segmentation, and compare it to \gls*{mbis}.
We also present the design considerations that underlie
  this work, and we describe the evaluation framework.
In \autoref{sec:results}, we describe the specific details of each
  experiment, illustrating the usefulness of \gls*{mbis}
  and reporting the results of evaluation.
Finally, we discuss in \autoref{sec:discussion} the three 
  experiments, and envision the unique opportunity that
  multivariate segmentation of the latest \gls*{mri} sequences
  provides. 

\makeatletter{}\section{Computational methods and theory}
\label{sec:methods}

\subsection{Background}
\label{sec:background}

Mixture models allow the expression of relatively complex marginal distributions
  fitting the observed variables in terms of more tractable joint
  distributions over the expanded space of observed and latent variables
  \citep{bishop_pattern_2009}.
The latent variables behave as simpler components used for building
  the inferred distribution from the observed data.
This general statistical framework provides not only the possibility
  of modeling complex distributions,   but also enables data to be clustered, using Bayes' theorem.
Given the generation and reconstruction processes involved in brain
  \gls*{mri}, it is accepted that these latent variables (the tissue
  classes) are reasonably well modeled with normal distributions
  \citep{van_leemput_automated_1999-1}.
Nonetheless, the existence of other minor sources of tissue contrast
  and the non-normality of several tissues under some conditions is
  widely accepted.
For instance, the \gls*{csf} is usually modeled with more than one
  normal distribution \citep{van_leemput_automated_1999-1,
  ashburner_unified_2005} to overcome these drawbacks.

A second relevant assumption is that the multivariate
  distributions associated with each expected cluster do
  not significantly overlap.
In the case of \gls*{mri} data, there are two principal sources of
  overlap in the observed tissue distribution: the \gls*{pv} effect and
  the \emph{bias field}.
On one hand, the so-called \gls*{pv} effect is remarkably related to
  tomographic biomedical imaging. Given that the images are defined
  on a grid of volume elements (voxels), they enclose a finite region.
This region may contain a mixture of signals from several
  tissues, producing an overlap between the tails of their distributions
  that can make the problem intractable by means of a
  \gls*{multivariate_gaussian}.
The number of voxels affected by the \gls*{pv} effect within a typical
  \gls*{mri} volume is usually significant, and worse when the resolution
  is low \citep{bromiley_multi-dimensional_2008}.
Previous studies have dealt with \gls*{pv} using non-normal intensity distribution
  models for each tissue \citep{santago_statistical_1995,noe_partial_2001,
  tohka_fast_2004}, modeling each cluster with more than one normal distribution
  \citep{ashburner_unified_2005,cuadra_comparison_2005},
  modeling the \gls*{mri} relaxation times at \gls*{pv}-affected voxels
  \citep{duche_bi-exponential_2012}, or
  using models with continuous latent variables \citep{liang_em_2009}.

On the other hand, most imaging datasets are affected to some degree
  by a spatially smooth offset field (called \emph{bias field}).
In \gls*{mri}, this illumination artifact derives from the spatial
  inhomogeneity of the magnetic field inside the scanner during
  acquisition.
Some retrospective techniques for tackling the bias field have
  been proposed, either embedded within the model
  \citep{van_leemput_automated_1999} or as a
  preliminary process \citep{tustison_n4itk:_2010}.

Finally, as \glspl*{multivariate_gaussian} are very sensitive to
  noise.
It is possible to introduce piecewise smoothness including
  spatial information in the described model, often
  implemented as a \gls*{markov}.
  
\subsection{Distribution model}
\label{sec:model}

\paragraph{\Acrlong*{multivariate_gaussian}}
Let $Y = \{\vy_i \in \mathbb{R}^C \}$ be a random variable that
  represents the observed data. Therefore, the image $Y$ is a stack
  of $C$ different \gls*{mri} sequences, and $i \in [1,\ldots,N]$
  is the index of each voxel in this image of $N$ voxels.
Accordingly, segmentation aims to obtain a certain realization of
  the latent random variable $X = \{ x_i \}$.
Thus, $Y$ is segmented after finding the class identified by 
  $l_k$ in the set of $K$ different clusters 
  $\mathcal{L} = \{l_1, l_2, \ldots, l_K \}$ that best matches
  $\vy_i$ given the model.
Finally, the \gls*{multivariate_gaussian} model is defined by two
  probabilities.
The first is the estimated normal distribution of each
  cluster, $\mathcal{N}(\vy_i \mid \theta_k )$, with
  $\theta_k = \{\boldsymbol{\mu}_k,\boldsymbol{\Sigma}_k\}$
  the parameters (means vector and covariance matrix)
  corresponding to the tissue identified by label $l_k$.
The second is the \emph{prior} probability of every voxel $i$
  belonging to cluster $l_k$, represented by $\pi_{k,i}$.

Using Bayes' theorem and the multivariate normal distribution as
  starting points, segmentation relies on iteratively improving the
  fitness of the model to the data.
To this end, \emph{posterior density} or \emph{responsibility} maps
  can be computed to evaluate the fitness \citep{bishop_pattern_2009}
  using the following expression:
\begin{equation}
\label{eq:post_density}
\gamma_{k,i}=P(x_i = l_k \mid \vy_i) = \frac
{\pi_{k,i} \, \mathcal{N}(\vy_i \mid \theta_k )}
{\sum_{j\in K} \pi_{j,i} \, \mathcal{N}(\vy_i \mid \theta_j ) },
\end{equation}
where $\gamma_{k,i}$ is the \emph{posterior density} of tissue class 
  $k$ at voxel $i$. Equivalently, $\gamma_{k,i}$ is the probability of
  detecting the class $l_k$ at $i$, given that $\vy_i$ was observed
  and the current model defined by $\lbrace \pi_{k,i}, \theta_k \rbrace$.
  
Once a stopping criterion has been met, the \emph{fuzzy} segmentation
  outcome is the set of \glspl*{tpm} corresponding to the last $\gamma_{k,i}$
  estimated, and the \emph{hard} segmentation $X$ is obtained after
  applying the \gls*{map} rule:
\begin{equation}
\label{eq:map}
\tilde{x}_i = \underset{\mathcal{L}}{\textrm{argmax}} \lbrace \gamma_{k,i} \rbrace
\end{equation}

\paragraph{Correction for bias field}
\label{sec:bias_model}
Let $B = \lbrace \vec{b}_i \in \mathbb{R}^C \rbrace$ be the unknown bias field,
  with $C$ independent components (one per input \gls*{mri} sequence).
It is a widely accepted assumption to consider $B$ a multiplicative smooth
  function of the pixel position \citep{vovk_review_2007}. Thus,
  we introduce this new random variable on the definition of the observation
  $\vy_i = \hat{\vy}_i \cdot \vec{b}^{T}_i$, where $\hat{\vy}_i$
  is the bias-free feature vector in $i$.
  
In order to extract $\hat{\vy}_i$, the observed variables $\vy_{i}$
  are logarithm transformed, so that $B$ becomes an additive field.
Thus, $B$ can be estimated by fitting a smooth function that minimizes
  the error field $E = \lbrace \vec{e}_i \rbrace$:
\begin{equation}
\label{eq:bias_error}
\vec{e}_{i} = \log{\hat{\vy}_{i}} - \log{\sum_{k\in K} \gamma_{k,i}\,\boldsymbol{\mu}_{k}}.
\end{equation} 
In \autoref{sec:em}, we shall discuss how to introduce the minimization
  of $E$ into the optimization routine for the estimation of $B$.
  
\paragraph{Regularization}
\label{sec:hmrf}
Finally, spatial constraints are included within the model in
  order to obtain a piece-wise smooth and plausible segmentation.
Typically, \gls*{multivariate_gaussian} methods are combined with the
  \gls*{markov} model to introduce such regularization.
The origin of \glspl*{markov} theory is the Gibbs distribution
  \citep{geman_stochastic_1984}, which has been comprehensively
  covered in the literature \citep{li_markov_2009}.
The spatial constraints are induced in the model throughout the 
  \emph{proportion factors} $\pi_{k,i}$ \eqref{eq:post_density}.
Therefore, assuming an \gls*{markov} model, $\pi_{k,i}$ now varies
  depending on the tissues located at the neighboring sites of $i$,
  the so-called \emph{clique} $\mathcal{N}_{i}$,
  with $i \notin \mathcal{N}_{i}$ and
  $i \in \mathcal{N}_{j} \iff j \in \mathcal{N}_{i}$.
\begin{equation}
\label{eq:hmrf_energy}
\pi_{k,i} \propto e^{\left(V_{i}(x_{i}=l_k)+\frac{\lambda_{\mathcal{N}}}{2}\underset{j\in\mathcal{N}_{i}}{\sum}V_{ij}(x_{i}=l_k,x_{j})\right)},
\end{equation}
where $V_{i}(x_{i})$ is an external field that weights the relative
  importance of the different classes present in the image and $V_{ij}(x_{i},x_{j})$
  models the interactions between neighbors.
Generally, $V_{i}(x_{i})=0$ is set in order to use a simplified model.
A typical definition of $V_{ij}(x_{i},x_{j})$ follows Pott's model
  \citep{zhang_segmentation_2001}:
\begin{equation}
\label{eq:potts_model}
V_{ij}(x_{i},x_{j})=\delta(x_{i},x_{j})=\begin{cases}
1, & \textrm{if}\: x_{i}=x_{j}\\
0, & \textrm{otherwise}.
\end{cases}
\end{equation}

\subsection{Optimization}
\label{sec:optimization}
Typically, the most common optimization of the described model has been
  solved by the \gls*{e_m} algorithm.
With the inclusion of the \glspl*{markov} into the model, the problem
  turns out to be a combinational one, intractable with \gls*{e_m}.
Therefore, a second solver is usually required for optimization
  of the full model.
A number of algorithms have been proposed for this application 
  \citep{bishop_pattern_2009}, for instance \gls*{icm},
  \gls*{montecarlo} sampling, or \gls*{graph_cuts}.

\paragraph{\Acrlong*{e_m} algorithm}
\label{sec:em}
\Gls*{e_m} iteratively seeks local solutions that are
  constantly closer to the global one.
For further details, we refer the reader to a theory book
  \citep{bishop_pattern_2009}.
In \autoref{alg:e_m}, we describe a modified version including
  the bias model estimation.
The \gls*{e_m} algorithm requires a good initialization of
  $\lbrace \pi_{k,i}, \theta_k \rbrace$, as it is
  likely to get trapped in local minima.
Typical initialization strategies can be automated, as the k-means algorithm,
  or the application of prior knowledge using \glspl*{tpm} from an atlas
  to estimate the initial parameters.
In addition, manual initialization is possible, explicitly specifying the
  model parameters.

\paragraph{\Acrlong*{graph_cuts} optimization}
\label{sec:background_graph_cuts}
The standard optimization procedure is to approximate the solution
  with the \gls*{e_m} algorithm and then impose the \gls*{markov}
  implicit regularization, as depicted in \autoref{fig:em_flowchart}.
The problem is stated so that we seek the labeling $X$ 
  that minimizes the following energy functional \citep{boykov_fast_2001}:
\begin{equation}
\Phi(X,Y)=\Phi_{smooth}(X)+\Phi_{data}(X,Y)
\label{eq:gc_energy}
\end{equation}
where $\Phi_{smooth}$ reflects the extent to which $X$ is not
  piecewise smooth, while $\Phi_{data}$ measures the 
  disagreement between $X$ and the observed data $Y$.

\Gls*{graph_cuts} algorithms approximately minimize the energy $\Phi(X,Y)$
  for the arbitrary finite set of labels $\mathcal{L}$ under two fairly
  general classes of interaction penalty $V_{ij}$:
  \emph{metric} and \emph{semi-metric} \citep{boykov_fast_2001}.
In the case of $n=2$ this solution is exact, as opposed to greedy algorithms
  like the widely used \gls*{icm}.
Weighted graphs encoding all possible energy configurations
  are built as follows.
The nodes of the graph are the two possible labels and each voxel of
  the image grid.
All nodes corresponding to image voxels are linked to the nodes
  of the labels, encoding on the edge weight the membership likelihood.
Edges between voxel nodes encode the pair-wise interactions of the
  \gls*{markov} system.
The minimum of the energy functional \eqref{eq:gc_energy} concurs on 
  the minimum cut of the graph.
In graph theory, a cut is a partition of the vertices of the graph
  in disjoint subsets.
The size of a cut depends on the number and weights of the edges
  removed.
Therefore, the minimum cut is that not larger than the size of any
  other cut.
  
The binary case is extended to $n$-cluster classification
  with iterative algorithms of very large binary \emph{moves}
  (a simultaneous and large change of assigned labels in $X$).
The basic underlying concept is to find local minima sequentially
  at each iteration, based on the allowed moves.
\citeauthor{boykov_fast_2001} \citep{boykov_fast_2001,kolmogorov_what_2004}
  proposed two different algorithms to implement \gls*{graph_cuts},
  called $\alpha$-expansion and $\alpha\beta$-swap.
In \autoref{alg:swap} (Appendix B), we describe $\alpha\beta$-swap to illustrate
  how the iterative minimization works.
Both algorithms have been proven to be highly accurate and efficient
  approximations of the global minimum for $n$-cluster classification
  \citep{boykov_experimental_2004}.
  
\subsection{Implemented methods and contributions}
\label{sec:implementation_details}
\begin{figure*}[t]
	\includegraphics[width=1.0\linewidth]{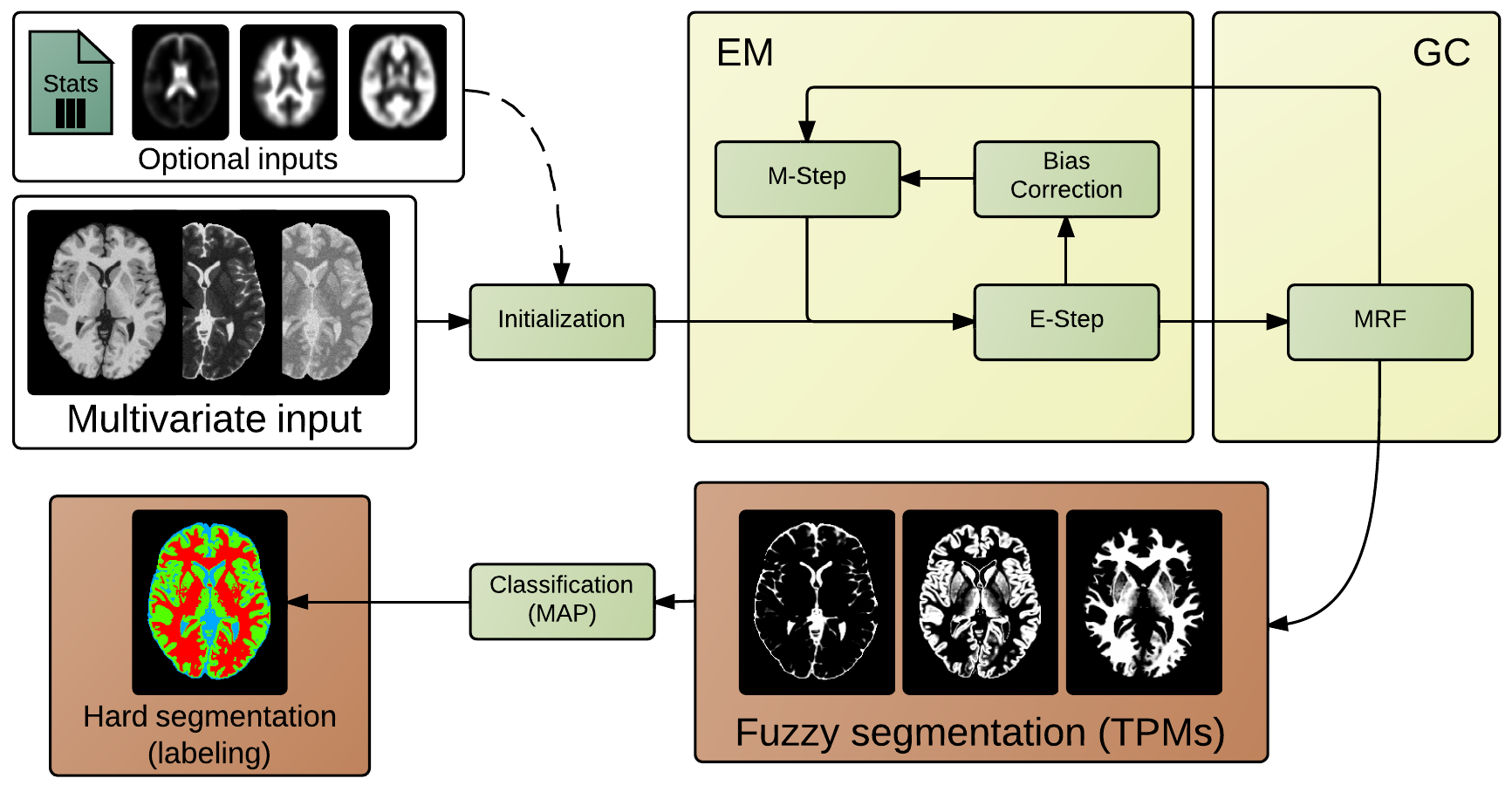}
	\caption[Segmentation flowchart]{\Gls*{e_m}-\gls*{graph_cuts} segmentation
	takes as inputs the blocks depicted in white background and produces the
	blocks in brown background as outputs.
	Typically, the initialization can be performed supplying a file with
	the parameters of the model, or prior \glspl*{tpm} from an atlas (optional
	inputs are represented with a dashed line connector).}
	\label{fig:em_flowchart}
\end{figure*}

\Gls*{mbis} implements the general \gls*{multivariate_gaussian} model as
  described in \autoref{sec:model}.
We specify in this section the main contributions and features
  implemented in \gls*{mbis}.
An overview of the principal elements of the tool and the optimization
  strategy is presented in \autoref{fig:em_flowchart}.

\paragraph{Initialization}\label{par:initialization} Once the model has been fully defined (number of expected pure tissues,
  number of normal distributions per tissue, number of special \gls*{pv}
  classes, and bias correction), \gls*{mbis} allows
  for several standard initialization approaches.
One common and fully-automated strategy is the use of the
  k-means algorithm, which is the default option in \gls*{mbis} when
  no other initialization is required.
A second extended initialization strategy is manually setting $\{\theta_k\}$,
  assuming a uniform distribution for $\pi_{k,i}$.
Finally, it is also common to use atlas priors when the spatial mapping
  between the actual case and the atlas is known.
Atlas priors can be supplied to \gls*{mbis} as a set of \glspl*{tpm},
  one per normal distribution.
It is important to note that these priors are no longer applied
  after initialization.

\paragraph{Bias correction}\label{par:bias_correction} When bias correction is required, a new definition of likelihood
  derived from \eqref{eq:post_density} is applied.
We estimate the bias field $B$ approximating the error measurement map
  $E$ obtained after \eqref{eq:bias_error} with uniform B-splines.
This solution is dual to N4ITK, the non-parametric algorithm presented
  elsewhere \cite{tustison_n4itk:_2010}.
\citeauthor{tustison_n4itk:_2010} analyzed the best B-spline parametrization
  for bias correction, and concluded that it is preferable to other models
  based on linear combinations of polynomial or smooth basis functions.
Before the next iteration of the E-step (see \autoref{alg:e_m}), data are
  corrected with the field vector $\vec{b}_{i}$ at $i$ before the distribution 
  parameters are calculated.

\paragraph{\Acrlong*{pv} model}\label{par:model} On the basis of previous findings \citep{cuadra_comparison_2005}, \gls*{mbis}
  tackles the \gls*{pv} effect by modeling pure tissues with \emph{in-class}
  mixtures of normal distributions, and by adding specific \gls*{pv} classes
  \citep{noe_partial_2001}.
Appropriate transition penalties can be set consistently for these classes, as
  in \citep{cuadra_comparison_2005}.
Instead of estimating the tissue contributions to the \gls*{pv} classes
  within the model, we provide a simplified procedure to achieve this aim
  \emph{a posteriori}.
The methodology computes the Mahalanobis distance \eqref{eq:mahalanobis}
  of the \gls*{pv} samples to the tentative pure tissues.
Interpreting the posterior probability as a volume fraction of the tissue
  within the voxel, this volume is divided between the pure tissues inversely
  proportional to the distance $D_k$ \eqref{eq:mahalanobis} to the tissues.
This \gls*{pv} solving is applied to the experimental results presented
  in \autoref{sec:results}.
\begin{equation}
\label{eq:mahalanobis}
D_k(\vy_i) = \sqrt{(\vy_{i}-\boldsymbol{\mu}_{k})^{T}\boldsymbol{\Sigma}_{k}^{-1}(\vy_{i}-\boldsymbol{\mu}_{k})}.
\end{equation}

\paragraph{\Acrlong*{graph_cuts} optimization}\label{par:graphcuts_optimization} \Gls*{mbis} implements \gls*{graph_cuts} optimization as in \citep{boykov_fast_2001},
  wrapping the \emph{maxflow} library (\url{http://vision.csd.uwo.ca/code/})
  in \gls*{itk} to solve the graphs.
The weighting parameter $\lambda_{\mathcal{N}}$ \eqref{eq:hmrf_energy} must be
  adequately determined for sensible regularization.
In \autoref{sec:brainweb_evaluation}, we describe the experiment conducted to
  set $\lambda_{\mathcal{N}}$ empirically.
The special \gls*{pv} classes are taken into account specifying an appropriate
  transition model.
The transition model is a matrix where the interactions between individual normal
  distributions are defined.
These energy interactions are defined by the $V_{ij}(x_{i},x_{j})$ presented in
  \autoref{sec:hmrf}.
Generally, in an \gls*{markov} model including several normals distributions per
  tissue (to account for \gls*{pv} effects), transitions within pure tissue have
  lower penalties (inner transitions) than transitions between pure tissues
  (outer transitions).
\Gls*{mbis} supports complex neighboring systems (beyond the simplest Pott's model
  \eqref{eq:potts_model}), distance weighted energy interactions, and non-metric
  tissue transition models. 

\makeatletter{}\section{Software description}\label{sec:software}

\subsection{Existing software}
\label{sec:related_work}

Many fully automated brain tissue segmentation tools, based on 
  Bayesian classifiers, are readily available and widely used.
In \autoref{table:tools_comparison}, we present a comparison among
  representative existing tools, along with a brief summary of the
  unique features of each.
All the tools make use of the \gls*{multivariate_gaussian} model
  with \gls*{markov} regularization.
The tools listed in the table are \gls*{fast} \citep{zhang_segmentation_2001}, 
  \gls*{spm}, \gls*{leemput_ems}, \emph{ATROPOS} \citep{avants_open_2011},
  \emph{NiftySeg} \citep{cardoso_niftyseg:_2012}, \emph{Freesurfer} 
  \citep{fischl_freesurfer_2012},
  and the software proposed in the present study (\gls*{mbis}).
The presented tools generally share a base design that follows
  the flowchart in \autoref{fig:em_flowchart}.
It is important to note that Freesurfer and \gls*{spm} are not just
  segmentation utilities, but fully automated pipelines for brain
  \gls*{mri} processing and analysis that include brain tissue segmentation.
\Gls*{spm} provides an isolated interface (called \texttt{segment}) for the
  problem at hand, the methodology of which is described elsewhere
  \citep{ashburner_unified_2005}.
Conversely, Freesurfer provides precise \emph{hard} segmentations of the brain
  in a large number of individual neuroanatomical regions \citep{fischl_whole_2002},
  which can be appropriately fused to the three-tissue problem.
The features presented in \autoref{table:tools_comparison} regarding Freesurfer
  and \gls*{spm} refer only to their whole-brain segmentation processes.

\begin{table*}[!t]
\begin{minipage}[t]{\linewidth}
\caption[Brain tissue segmentation tools]{\label{table:tools_comparison}}
\rowcolors{2}{white}{lightgray}
\footnotesize
\makeatletter{}\begin{tabularx}{\textwidth}{l|ZZZZZZ||Z}
\hline
              & \acrshort*{fast} & \acrshort*{spm} & \acrshort*{leemput_ems}     & ATROPOS    & NiftySeg & Freesurfer & \gls*{mbis} \\
\hline
Multivariate  & Partial    & Partial$^{1}$ & Full       & Full      &  Full      & No  & Full \\
Optimization  & \acrshort*{icm}  & \acrshort*{icm} & \acrshort*{montecarlo}& \acrshort*{icm}  & Unknown   & \acrshort*{icm} & \acrshort*{graph_cuts} \\
Bias model    & Polynomial & \acrshort*{dct} & Polynomial & No$^{3}$  & Unknown & No$^{3}$ & B-spline \\
Atlas usage   & Available  & Intensive & Available  & Available &  Intensive & Intensive & Available \\
License       & GPL        & GPL       & BSD-like   & BSD       &  BSD       & Freeware & GPL \\
Platform      & Unix       & Matlab    & SPM8       & Any$^{2}$       &  Any$^{2}$       & Unix & Any$^{2}$ \\
Reference     & \citep{zhang_segmentation_2001} & \citep{ashburner_unified_2005} & \citep{van_leemput_unifying_2003} & \citep{avants_open_2011} & \citep{cardoso_niftyseg:_2012} & 
\citep{fischl_whole_2002} & \\
\hline
\end{tabularx} 
\footnotesize
$^{1}$ Work in progress. \\
$^{2}$ \emph{Any} platform supported by the \emph{CMake} building system.\\
$^{3}$ The tool does not integrate a bias model, but it is released along with an external tool for correction.\\
\end{minipage}
\end{table*}

The first feature to be compared is multivariate implementation.
\Gls*{leemput_ems}, ATROPOS and NiftySeg fully support the
\gls*{multivariate_gaussian} model.
\Acrshort*{spm} is currently integrating support for multivariate
  data, while \gls*{fast} provides multichannel segmentation that
  importantly differs from the univariate segmentation methodology.
Freesurfer only supports \gls*{t1} \gls*{mri} as input for segmentation.

The model estimation is always performed with the \gls*{e_m} algorithm,
  possibly with some improvements.
For instance, \gls*{leemput_ems} implements a
  robust estimator and \gls*{pv} constraints.
Therefore, this property has been omitted in \autoref{table:tools_comparison}.
The main differences are found in the \gls*{markov} energy minimization 
  problem, \gls*{icm} being the most used methodology.
\Gls*{leemput_ems} implements \acrfull*{montecarlo} sampling, which is more
  reliable than \gls*{icm} but computationally expensive.
\Gls*{mbis} is the first tool among the surveyed software packages to include
  \gls*{graph_cuts} optimization, for which a great trade-off
  between efficiency and correctness has been proven.

Another important feature is the bias field correction, generally solved by
  approximation of linear combinations of smooth basis functions.
\Gls*{fast} and \gls*{leemput_ems} use polynomial least-squares fitting,
  \gls*{spm} uses the \gls*{dct} with \gls*{lm} optimization, and \gls*{mbis}
  uses B-splines basis.
Unfortunately, there was no information available about the bias model
  implemented in NiftySeg at the time of writing.
Two of the surveyed tools do not internally integrate a bias model:
Freesurfer provides a pipeline including a previous correction utility,
  and ATROPOS advises the prior use N4ITK \citep{tustison_n4itk:_2010}.

The next point of comparison is the use of atlases to initialize the algorithm and/or
  to aid the estimation of model parameters.
All the tools can initialize segmentation using prior atlas information.
Those tools that also use priors throughout the model fitting are denoted
  with ``intensive'' atlas use in \autoref{table:tools_comparison}.

In terms of software availability, for all the tools the source
  code is publicly released and the software is distributed under 
  open-source licenses.
With respect to their installation, a number of them (\gls*{fast}, \gls*{spm} 
  and \gls*{leemput_ems}) are platform-dependent, whereas the others
  are multi-platform using the \emph{CMake} building tool (\url{http://www.cmake.org}).

\subsection{Design considerations}\label{sec:design}
Given the described context of existing software, we aimed to design a
  multivariate segmentation tool, which is flexible, easy to use, comprehensive,
  and would also include a \gls*{graph_cuts} solver and a B-spline bias field model.
As a result we designed \gls*{mbis}, an open-source and cross-platform software that
  supports multivariate data by design and that integrates all the methods described in
  \autoref{sec:implementation_details}.
Segmentation provided with \gls*{mbis} is general purpose.
In this study, \gls*{mbis} is specifically adapted to the 3D
  brain tissue segmentation problem.
In order to facilitate contributions by third-party developers, the code
  follows the standards of \gls*{itk}, and some interfaces have been defined
  to integrate new code, preserving the software modularity.

We also promote \emph{reproducible research}, a concept that is drawing increasing
  interest in parallel to the proliferation of computational solutions
  for image processing problems.
Following the definition of \citeauthor{vandewalle_reproducible_2009}
  \citep{vandewalle_reproducible_2009}, we release here an
  open-source bundle with evaluation experiments based on open data
  to help the community replicate and test our work \citep{yoo_open_2005,
  ibanez_open_2006}.
  
In order to evaluate \gls*{mbis} comprehensively, we define three
  validation targets.
Consistently with the design considerations mentioned above,
  we test the performance of \gls*{mbis} with three different
  open data resources containing multivariate and multi-site
  data.
Full details of these databases are provided in \autoref{table:data} 
  (\ref{a:data}), describing the \gls*{mri} sequences involved and
  their specific parameters.
Finally, we address these targets in three different experiments
  (the results are presented in \autoref{sec:results}).

The first experiment evaluates the accuracy of segmentation, with 
  comparison to \gls*{fast}, using one simulated dataset.
The evaluation framework includes tests to calibrate the best
  parameters for the tool, benchmarks of the bias field estimation,
  and studies the impact of spatial misalignment between channels.
The second experiment evaluates the reproducibility of results, in
  similar settings to a recent validation study \citep{de_boer_accuracy_2010}.
Unfortunately, the database used by \citeauthor{de_boer_accuracy_2010} is not
  publicly available.
Hence, the resulting figures are not directly comparable to their work as
  we used different data.
On one hand, we studied the repeatability of the segmentation by analyzing
  the differences in tissue volumes.
On the other hand, the overlap indices described in \autoref{sec:evaluation_indices}
  were evaluated.
The third section of the evaluation framework is an exemplary pipeline of
  tissue volume analysis in large-scale databases.
We illustrated the use of \gls*{mbis} on such applications, segmenting
  multivariate \gls*{mri} datasets from 584 healthy subjects, and correlating
  tissue volumes with age.

\subsection{Evaluation framework}
\label{sec:experimental_framework}
The evaluation framework is built using \gls*{nipype}, in order to facilitate the
  fulfillment of the requirements of reproducible research.
The evaluation includes a \gls*{nipype} \emph{Interface} to \gls*{mbis},
  three \gls*{nipype} \emph{Workflows} to implement the experiments
  described in \autoref{sec:design} and a set of scripts in Python to automate the
  execution of the workflows and presentation of results (figures and tables
  included in this paper).
To assess and compare results appropriately in terms of accuracy and 
  robustness \citep{altman_measurement_1983}, we evaluate two families of
  indicators: volume agreements and overlap indices.
  
\paragraph{Volume agreement}
Volume agreement between the segmentation found and the ground-truth, or
  between segmentations of corresponding time points, is a commonly used
  benchmark.
Volumes of the identified tissues are directly related to the total size
  of the brain.
Therefore, we provide here the ``\gls*{icv} fraction'' of each tissue
  as the ratio of the measured volume over the total volume of
  the whole-brain.

\paragraph{Overlap indices\label{sec:evaluation_indices}}
Overlap is a widely used indicator to assess segmentation results
  with respect to a ground-truth \citep{crum_generalized_2006}.
We use a \gls*{ev_fsi} derived from the fuzzy \gls*{ev_ji} \citep{crum_generalized_2006},
  as in eq. \eqref{eq:si_index}.
This fuzzy index definition takes the resulting \glspl*{tpm} as inputs,
  and naturally extends the binary definition.
We refer to the binary index as the \gls*{ev_si}, when computed on
  \emph{hard} segmentations.
Generally, the definition of \gls*{ev_ji} tends to favor classes with greater volume
  when computing the average overlap of several classes.
Thus, results reporting averages of overlap indices are compensated
  for tissue volume in this paper.
Additional indices are also provided:
  \gls*{ev_tpf} that acts as a measure of sensitivity \eqref{eq:tpf_index};
  \gls*{ev_ef} that expresses the over-segmentation \eqref{eq:ef_index};
  and \gls*{ev_oc}, which is reported as an alternative to the 
  \gls*{ev_si} \eqref{eq:oc_index}.
\begin{align}
fSI,SI&=\frac{2\,JI}{1+JI}\,, \label{eq:si_index} \\
TPF&=\frac{TP}{TP+FN}\,, \label{eq:tpf_index} \\
EF&=\frac{FP}{TP+FN}\,, \label{eq:ef_index} \\
OC&=1-\frac{FP+FN}{TP}\,, \label{eq:oc_index}
\end{align}
where $TP$ stands for true positives, $FP$ for false positives, and $FN$ for false 
negatives. We did not extend these measures to the probabilistic results, so they
are only used for the assessment of the \emph{hard} segmentations.

\makeatletter{}\section{Evaluation experiments and results}
\label{sec:results}

\subsection{Accuracy assessment and bias field correction}
\label{sec:brainweb_evaluation}

\begin{figure}
	\centering
	\includegraphics[width=1.0\linewidth]{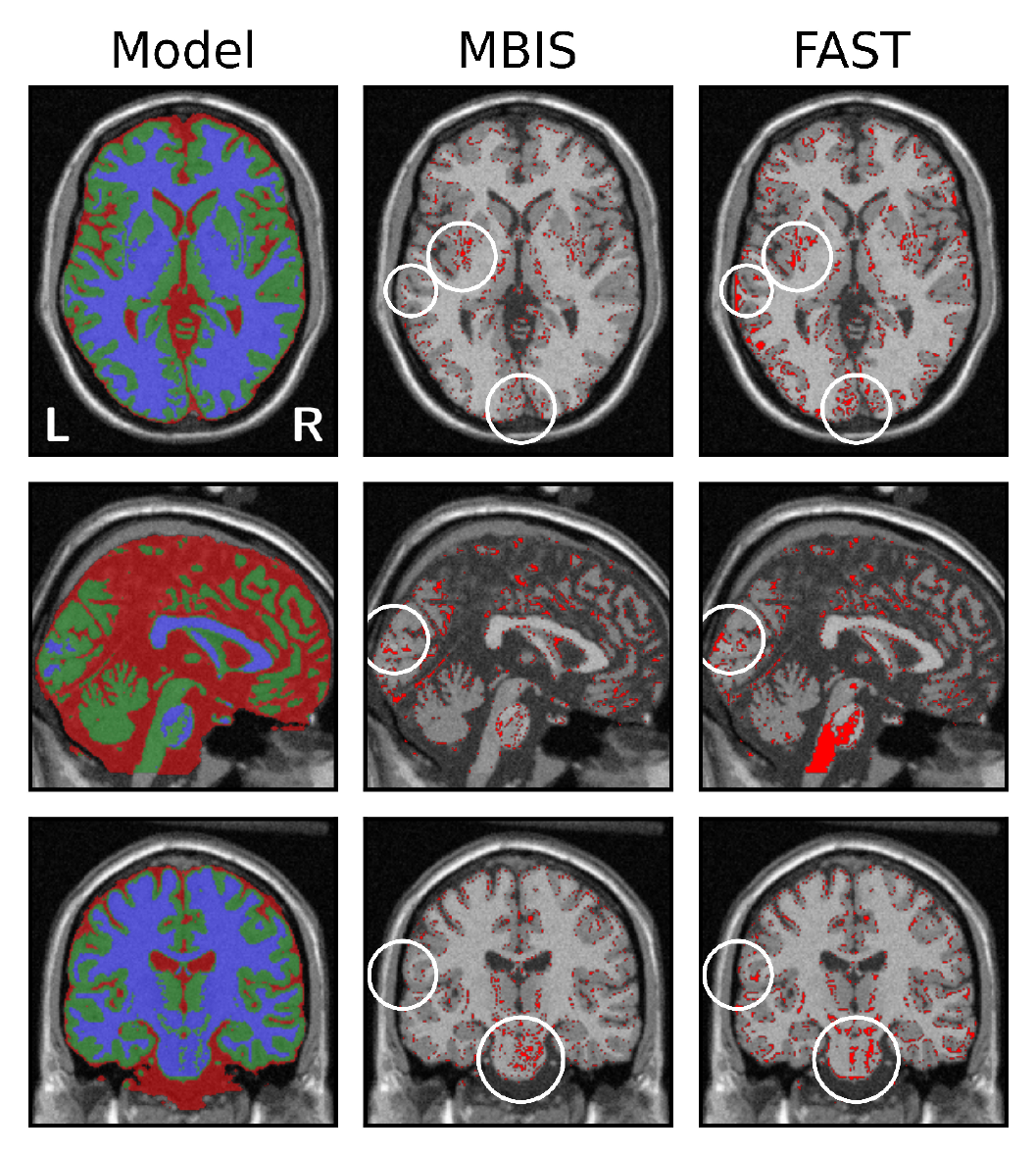}
	\caption[Visual assessment of the hard-segmentation results]{The ground-truth is presented on the left column.
	Error maps are presented for \gls*{mbis} and \gls*{fast} (red color represents misclassified pixels).
	Rather than the obvious differences, we highlighted some more subtle examples.
    \label{fig:brainweb_hard}}
\end{figure}

\paragraph{Data}
The first experiment demonstrated the accuracy of \gls*{mbis}, and it was conducted
  over the only multivariate dataset included in the \emph{BrainWeb Simulated 
  Brain Database} \citep{cocosco_brainweb:_1997}.
The details of this dataset can found in \autoref{table:data}.
It provides \gls*{t1}, \gls*{t2} and \gls*{pd} realistic
  \gls*{mri}, simulated for only one ``normal'' model presenting a healthy anatomy.
We generated the ground-truth taking the \glspl*{tpm} corresponding to soft tissues
  from the BrainWeb (i.e. \gls*{csf}, \gls*{gm}, glial matter, and \gls*{wm}).
As glial matter and \gls*{gm} are almost indistinguishable (in terms of intensity)
  for the three \gls*{mri} (\gls*{t1}, \gls*{t2} and \gls*{pd}), we merged their
  corresponding \glspl*{tpm} in order to produce a three-class (\gls*{csf}, \gls*{gm},
  and \gls*{wm}) distribution model.
The resulting three \glspl*{tpm} were normalized to sum up to 1.0 at every voxel.
Using the \gls*{map} criterion \eqref{eq:map}, we generated the ground-truth labeling
  (presented in the first row of \autoref{fig:brainweb_hard}).
As it is shown in the same figure, the combination of the three \glspl*{tpm} yields a
  brain mask that was used for brain extraction.

\paragraph{Parameter $\lambda_{\mathcal{N}}$ setting}
We characterized the parameter $\lambda_{\mathcal{N}}$ \eqref{eq:hmrf_energy},
  to set an appropriate default value.
We used the BrainWeb dataset with a noise factor of 5\% and the three available
  inhomogeneity fields (called 0\%, 20\% and 40\% with increasing 
  strength of intensity non-uniformity field).
\begin{figure*}
	\includegraphics[width=1.0\textwidth]{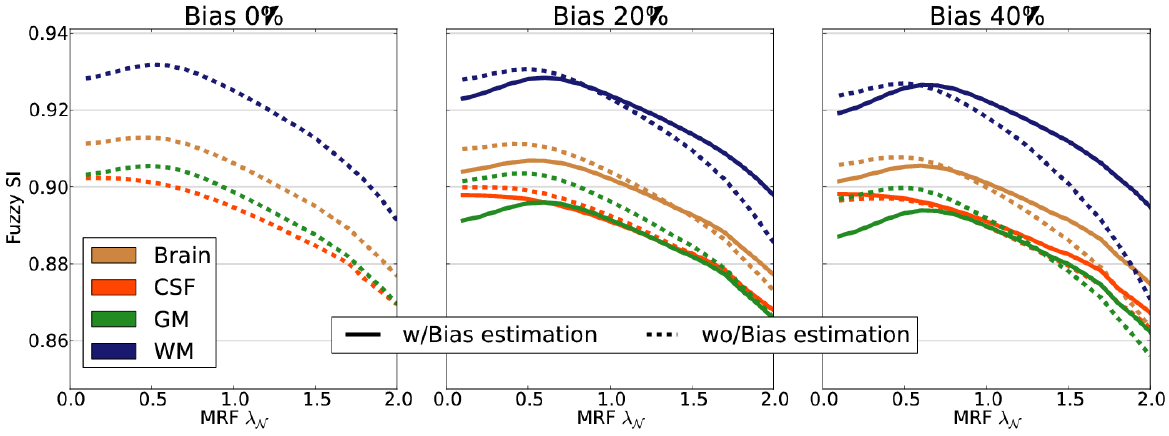}
	\caption{\textbf{Parameter calibration.} Fine tuning of the parameter $\lambda_{\mathcal{N}}$ 
	\eqref{eq:hmrf_energy}. Dashed lines represent the evaluation without bias estimation and 
	filled lines correspond to the evaluation with bias estimation. The figure shows that the
	optimum value for $\lambda_{\mathcal{N}}$ ranges [$0.55-0.65$].
	}
	\label{fig:mrf_lambda}
\end{figure*}  
From the results shown in \autoref{fig:mrf_lambda}, two conclusions
  can be drawn.
First, $\lambda_\mathcal{N}=[0.55-0.65]$ is consistently the optimum value;
  and second, the bias estimation does not effectively improve the segmentation
  results.
As the channels have independent inhomogeneity patterns, the model is less
  prone to this confounding effect, allowing more flexible \gls*{multivariate_gaussian}
  models without losing sensitivity.
This conclusion is confirmed later, on the visual assessment of the
  estimated bias field maps.
Once $\lambda_\mathcal{N}$ was set, we explored different \gls*{pv} models
  to segment the dataset.

\paragraph{Results}The performance test on the synthetic ground-truth was carried out on
  the three available relaxation-time-weighted sequences
  (\gls*{t1}, \gls*{t2}, and \gls*{pd}), with 5\% noise
  and 20\% bias field.
We configured \gls*{mbis} for fully automatic initialization (k-means)
  and $\lambda_\mathcal{N}=0.6$.
Additionally, \gls*{fast} was also used to perform the segmentation using
  its multichannel mode and default settings.
After evaluating numerous configurations, we achieved acceptable results 
  from \gls*{fast} with a four-class model as suggested elsewhere
  \citep{he_generalized_2008}.
We merged the \gls*{tpm} of the fourth class into the one corresponding to \gls*{csf}.
This ad hoc decision was taken after ensuring that the accuracy figures were the
  best we could reach using \gls*{fast}.
The quantitative results shown in \autoref{table:brainweb} indicate a better overall
  performance (row labeled as ``Brain'') of \gls*{mbis} for all the evaluated indices.
\begin{table}[!htbp]
\caption[Quantitative results of the accuracy assessment]{Overlap measured using
  the different indices proposed in \autoref{sec:evaluation_indices}. Boldface font
  highlights the best score for both tools at each overlap index.
  Row labeled as ``Brain'' represents the volume-corrected average of 
  the three detected clusters (\gls*{csf}, \gls*{gm}, \gls*{wm}). Columns contain
  the different indices evaluated (see \autoref{sec:evaluation_indices}):
  \acrfull*{ev_fsi}, \acrfull*{ev_si}, \acrfull*{ev_tpf}, \acrfull*{ev_ef}, and \acrfull*{ev_oc}.
  \label{table:brainweb}}
  \rowcolors{2}{white}{lightgray}
  \makeatletter{}\begin{tabularx}{1.0\linewidth}{Z|Z|Z|ZZZZ}
\hline
                          &             &  \gls*{ev_fsi}       &   \gls*{ev_si} &  \gls*{ev_tpf} &   \gls*{ev_ef}  &    \gls*{ev_oc} \\ 
\hline  
{\cellcolor{white}}       & \gls*{fast}   &  0.846          &  0.874          &           0.907 &            0.163 &            0.710 \\
\multirow{-2}{*}{Brain}   & \gls*{mbis}   &  \textbf{0.912} &  \textbf{0.940} &  \textbf{0.955} &   \textbf{0.079} &   \textbf{0.871} \\
\hline
{\cellcolor{white}}       & \gls*{fast}   &  0.863          &  0.889          &           0.993 &            0.241 &            0.750 \\
\multirow{-2}{*}{\gls*{csf}}& \gls*{mbis}  &  \textbf{0.900} &  \textbf{0.923} &  \textbf{0.997} &   \textbf{0.163} &   \textbf{0.834} \\
\hline
{\cellcolor{white}}       & \gls*{fast}   &  0.807          &  0.845          &           0.741 &   \textbf{0.014} &            0.632 \\
\multirow{-2}{*}{\gls*{gm}} & \gls*{mbis}  &  \textbf{0.904} &  \textbf{0.939} &  \textbf{0.923} &            0.043 &   \textbf{0.871} \\
\hline
{\cellcolor{white}}       & \gls*{fast}   &  0.868          &  0.888          &  \textbf{0.987} &            0.236 &            0.749 \\
\multirow{-2}{*}{\gls*{wm}} & \gls*{mbis}  &  \textbf{0.931} &  \textbf{0.956} &           0.945 &   \textbf{0.032} &   \textbf{0.908} \\
\hline
\end{tabularx} 
\end{table}

Qualitative evaluation using visual assessment and error maps is also reported.
\autoref{fig:brainweb_hard} presents representative views of the error maps
  obtained with the tools under comparison, highlighting regions with remarkable
  differences.
  
\begin{figure*}[!htbp]
	\centering
	\includegraphics[width=1.0\textwidth]{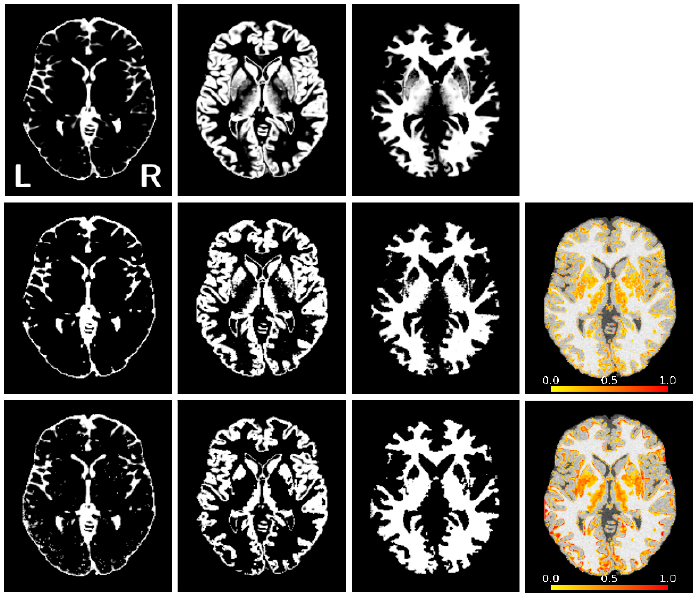}
	\caption[Visual assessment of the fuzzy-segmentation results]
	{First row shows the \glspl*{tpm} of the ground-truth (from left to right:
	\gls*{csf}, \gls*{gm}, \gls*{wm}). Second and third rows present the corresponding
	\glspl*{tpm} obtained with \gls*{mbis} and \gls*{fast}, respectively.
	The extra column represents the mean squared error of the \glspl*{tpm}, 
	normalized by the maximum squared error of both maps.}
	\label{fig:brainweb_fuzzy}
\end{figure*}

Regarding the \emph{fuzzy} outcome, \autoref{fig:brainweb_fuzzy} presents
  the \glspl*{tpm} obtained with \gls*{mbis} and \gls*{fast}, compared with
  the original ones.
An error map for each tool under testing is also presented, computed as the
  voxel-wise mean squared difference between the three original maps and
  the three maps obtained after segmentation.

\begin{figure}
	\centering
	\includegraphics[width=1.0\linewidth]{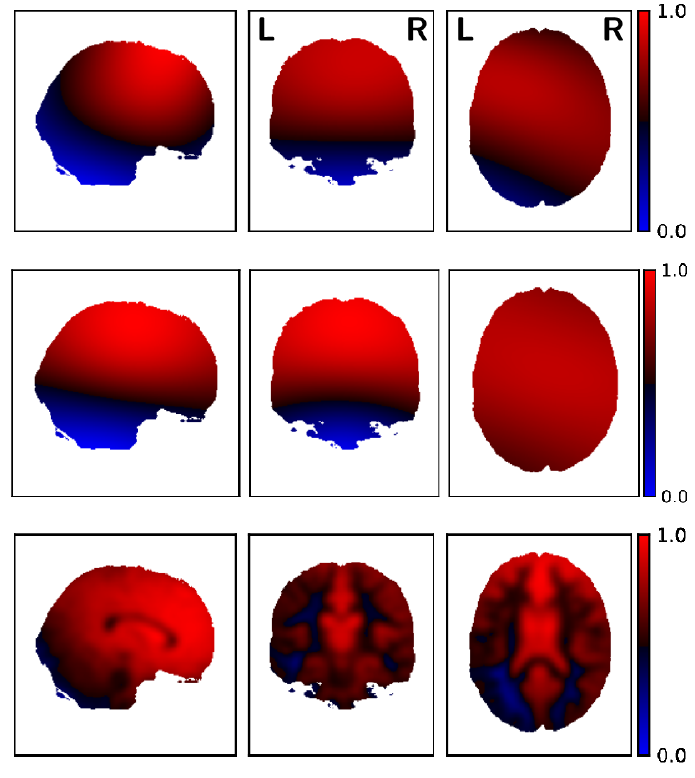}
	\caption[Comparison of bias estimation]
	{Normalized magnitudes of the reference (first row) and estimated field maps
	are presented (second row is \gls*{mbis}, third is \gls*{fast}),
	for the \gls*{t1} channel.}
	\label{fig:brainweb_results_bias}
\end{figure}

\paragraph{Bias field estimation}
We conclude the accuracy assessment by studying the performance on estimating
  the bias field.
\autoref{fig:brainweb_results_bias} presents a comparison of the results.
The first row shows the bias field contained by the simulated data from BrainWeb,
  for the \gls*{t1} \gls*{mri}.
The corresponding realizations of bias field for the \gls*{t2} and \gls*{pd}
  images are also available.
The second and third rows present the corresponding estimations obtained with 
  \gls*{mbis} and \gls*{fast}.
Visual assessment is straightforward, as \gls*{fast} did not perform a valid
  estimation of the bias field.
Similar results were obtained for the bias field that affected the \gls*{t2} 
  and \gls*{pd} images.
Even though \gls*{fast} obtained inadequate estimations, segmentation did not
  lose sensitivity dramatically (see \autoref{table:brainweb}), confirming
  that multivariate data are very robust against the different realizations of
  bias field on each channel, as they are independent.

\begin{figure}
  	\includegraphics[width=0.95\linewidth]{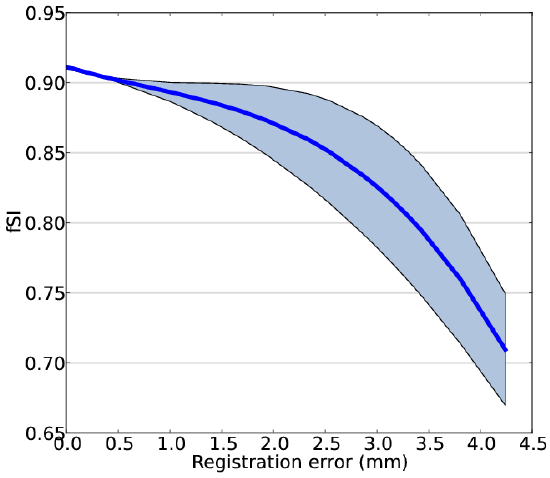}
    \caption[Intra-subject registration error]{
    	Influence of different combinations of translations
    	between the channels. The figure represents the \acrfull*{ev_fsi}
    	(defined in \autoref{sec:experimental_framework}) versus the absolute
    	displacement.
    	}
    \label{fig:segmentation_pitfalls}
\end{figure}

\paragraph{Intra-scan registration} The misregistration between the different contrasts stacked as a multivariate image
  is a prominent drawback that hinders multivariate segmentation.
We present in \autoref{fig:segmentation_pitfalls} the characterization of
  the impact of small misalignments between image channels.
More precisely, we translated \gls*{t2} and \gls*{pd} images from their 
  ground-truth location and conducted multivariate segmentation with \gls*{mbis}.
Segmentation results were assessed using the \gls*{ev_fsi} index, and they were
  proven to be quite sensitive to the registration error introduced artificially.
Given that we restricted the analysis only to 3D translations along the Y-axis, a very
  important impact should be expected from other misalignments (as rotations, linear
  transforms of a higher degree, or nonlinear deformations).

\subsection{Reproducibility evaluation}
\label{sec:kirby21}
\paragraph{Data}
The \emph{Multi-modal \gls*{mri} Reproducibility Resource}
  (also called the \emph{Kirby21 database}) \cite{landman_multi-parametric_2011}
  consists of scan-rescan imaging sessions on 21 healthy volunteers with no history
  of neurological disease.
The database includes a wide range of \gls*{mri} sequences, from which we selected
  \gls*{t1}, \gls*{t2} and \gls*{mt} for segmentation.
The complete database is publicly available online, and details of the \gls*{mri}
  sequences and other information can be found in \autoref{table:data}.

\paragraph{Image preprocessing}
First, all datasets were corrected for inhomogeneity artifacts using \gls*{n4itk} as
  it was necessary to obtain acceptable brain extraction using \gls*{bet}.
Moreover, the use of corrected images as input enabled testing the fully automated
  initialization included in \gls*{mbis}, avoiding the use of atlas information.
\Gls*{t1} images were then enhanced, replacing intensity values
  above the 85$^{th}$ percentile with the local median value.
This filtering removed the typical tail present in the intensity distribution of
  brain-extracted \gls*{t1} images, corresponding to spurious regions remaining
  after skull-stripping.
The second step, after this initial preparation, consisted of correctly aligning
  the different modalities with respect to the reference \gls*{t1} image.
We used \gls*{ants} to register rigidly the \gls*{t2} and \gls*{mt} images to the
  space of the \gls*{t1}.
We visually validated the intra-scan registration of each dataset, as it was
   proven to be an important source of error hindering repeatability in a previous
   experiment (\autoref{sec:brainweb_evaluation}).
   
\paragraph{Segmentation}We then used \gls*{mbis} and \gls*{fast} to segment the available datasets
  (a total of 42 datasets from 21 subjects scanned twice), using as input several
  variations of the three available \gls*{mri} sequences (i.e. \gls*{t1}, \gls*{t2},
  and \gls*{mt}).
We do not present a comparison of the repeatability with \gls*{fast} as most of the 
  resulting segmentations from it were not visually acceptable.
Even when the results were visually acceptable, they were not repeatable
  because of the well-known \emph{identifiability} problem \citep{bishop_pattern_2009}.
This problem occurs when a class is correctly detected, but assigned to a different 
  class-identifier, which makes the automatic computation of the evaluation indices
  impossible.
We performed segmentation using \gls*{mbis} with four different combinations of 
  sequences: \gls*{t1} alone, \gls*{t1}-\gls*{t2}, \gls*{t1}-\gls*{mt}, and 
  \gls*{t1}-\gls*{t2}-\gls*{mt}.
The first evaluation considering only the \gls*{t1} channel is the standard
  methodology and reference.
All segmentation trials used a five-class model, where four represented pure 
  tissues (two for \gls*{csf} and one each for \gls*{gm} and \gls*{wm}).
The remaining class fitted the partial volume existing between \gls*{csf} and \gls*{gm}.
We post-processed the \gls*{mbis} results to obtain the probability maps corresponding
  to three-class clustering, as described in \autoref{sec:implementation_details}.

\begin{figure*}[!htbp]
  \centering
  \subfigure[Volume agreement.]{  \label{fig:rep_volume}  \includegraphics[width=1.0\linewidth]{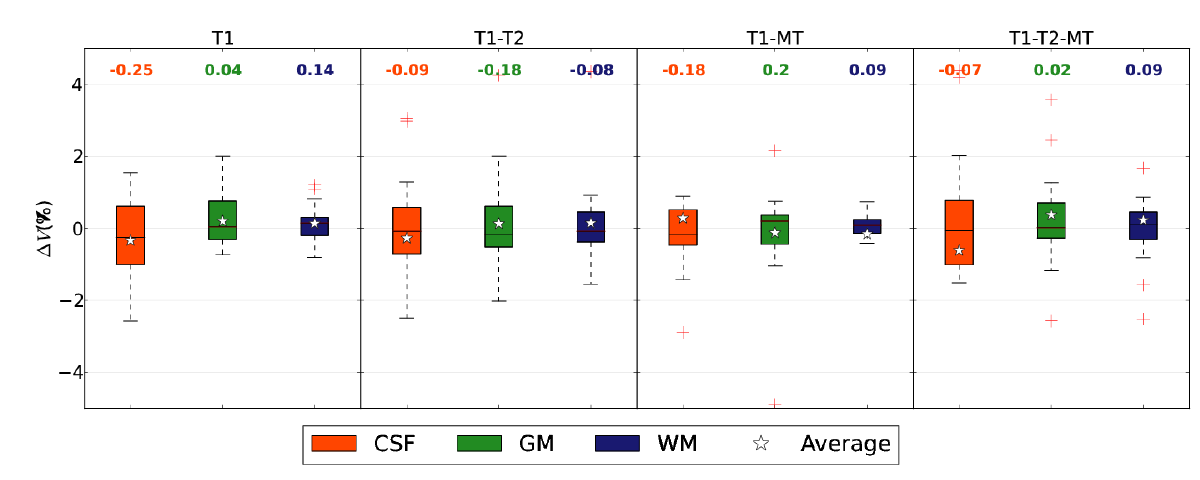}
  }\\
  \subfigure[Fuzzy Similarity Index.]{  \label{fig:rep_overlap}  \includegraphics[width=1.0\linewidth]{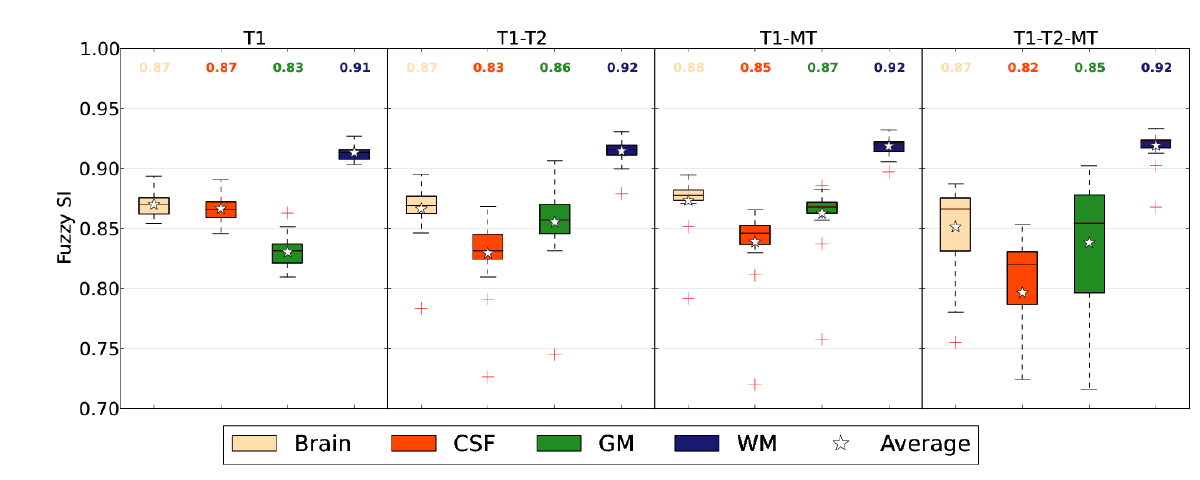}
  }
  \caption[Repeatability experiment]{Above each subplot, the \gls*{mri} sequences that
  were stacked to conduct each segmentation are indicated in the title.
  For instance, ``T1'' stands for \gls*{t1} alone, ``T1-T2'' stands for 
  \gls*{t1} and \gls*{t2}, and so forth.
  Inside the plots, the median value of each box is on top in the color of the corresponding tissue.
  \label{fig:repeatability}}
\end{figure*}

\paragraph{Results}
The first experiment consisted of measuring the volume change of each tissue 
 ($\Delta V_{\gls*{csf}}$,$\Delta V_{\gls*{gm}}$,$\Delta V_{\gls*{wm}}$) between the 
 two time points available for each subject (\autoref{fig:rep_volume}).
Bivariate approaches (\gls*{t1}-\gls*{t2} and \gls*{t1}-\gls*{mt}) decreased the volume
  agreement variance, thus providing a more robust outcome than using \gls*{t1} alone.
For the segmentation with three \gls*{mri} components, results were in the same range
  as for \gls*{t1} alone, but presented greater variance.
However, median tissue increments were closer to zero than the 
  monospectral segmentation medians.
The increased variance of results and the appearance of some additional
  outliers when segmenting in multichannel mode may be explained by two
  factors: a) the low quality of some datasets, where motion and gradient
  amplifier failure artifacts were present, and b) the misregistration of data
  between channels.
On one hand, some datasets on the \emph{Kirby21} database presented rather low quality,
  especially some of the \gls*{t2} images.
On the other hand, even though we visually validated intra-scan registration performance,
  some datasets were imperfectly aligned.

The second evaluation consisted of estimating the indices described
  in \autoref{sec:evaluation_indices}, after performing the segmentation
  independently over all of the two scan sessions and the four variations of
  multivariate inputs.
In order to measure the overlap between segmentations of the two scan sessions,
  an ``inter-scan'' alignment was performed by registering the \gls*{t1} image
  of the second scan to the first one with \gls*{flirt}.
The transform was used to resample the segmentation of the second scan input set
  in the space of the corresponding first scan.
\begin{table*}[!htbp]
   \caption[Quantitative results for overlap repeatability experiment]{All these measurements 
   are complementary to the results presented in \autoref{fig:rep_overlap}, and they are 
   computed with the hard segmentation results.
   First column indicates the tissue evaluated, labeling as ``Brain'' the weighted average
   of the other three.
   Second column specifies the \gls*{mri} sequences that were stacked as multivariate input
   (e.g. ``\gls*{t1}-\gls*{mt}'' means that the input feature vector contains samples drawn
   from the \gls*{t1} image as first component and from the \gls*{mt} image for the second).
   Remaining columns contain the different indices evaluated (see \autoref{sec:evaluation_indices}):
   \acrfull*{ev_si}, \acrfull*{ev_tpf}, \acrfull*{ev_ef}, and \acrfull*{ev_oc}.\label{table:rep_overlap}}
   \centering\footnotesize\rowcolors{2}{white}{lightgray}
   \makeatletter{}\begin{tabularx}{1.0\linewidth}{l|l|ZZZZ}
\hline
 &            &          \gls*{ev_si} &         \gls*{ev_tpf} &           \gls*{ev_ef} &            \gls*{ev_oc} \\
\hline 
{\cellcolor{white}}&\gls*{t1}&0.834$\pm$0.011&0.852$\pm$0.010&\textbf{0.192$\pm$0.016}&0.586$\pm$0.036\\
                    &\gls*{t1}-\gls*{t2}&\textbf{0.851$\pm$0.031}&\textbf{0.883$\pm$0.029}&0.207$\pm$0.051&\textbf{0.624$\pm$0.102}\\
{\cellcolor{white}}&\gls*{t1}-\gls*{mt}&0.840$\pm$0.034&0.873$\pm$0.032&0.221$\pm$0.061&0.590$\pm$0.115\\
\multirow{-4}{*}{Brain}&\gls*{t1}-\gls*{t2}-\gls*{mt}&0.847$\pm$0.036&0.877$\pm$0.037&0.203$\pm$0.044&0.602$\pm$0.134\\
\hline
{\cellcolor{white}}&\gls*{t1}&\textbf{0.800$\pm$0.016}&0.851$\pm$0.016&\textbf{0.277$\pm$0.040}&\textbf{0.499$\pm$0.052}\\
                    &\gls*{t1}-\gls*{t2}&0.754$\pm$0.042&\textbf{0.863$\pm$0.050}&0.434$\pm$0.168&0.339$\pm$0.161\\
{\cellcolor{white}}&\gls*{t1}-\gls*{mt}&0.746$\pm$0.047&0.858$\pm$0.075&0.452$\pm$0.207&0.307$\pm$0.183\\
\multirow{-4}{*}{CSF}&\gls*{t1}-\gls*{t2}-\gls*{mt}&0.739$\pm$0.064&0.814$\pm$0.091&0.388$\pm$0.117&0.269$\pm$0.279\\
\hline
{\cellcolor{white}}&\gls*{t1}&0.771$\pm$0.018&0.773$\pm$0.024&0.233$\pm$0.033&0.404$\pm$0.062\\
                    &\gls*{t1}-\gls*{t2}&\textbf{0.861$\pm$0.051}&0.853$\pm$0.070&\textbf{0.128$\pm$0.064}&\textbf{0.668$\pm$0.150}\\
{\cellcolor{white}}&\gls*{t1}-\gls*{mt}&0.835$\pm$0.057&0.830$\pm$0.080&0.159$\pm$0.105&0.593$\pm$0.168\\
\multirow{-4}{*}{GM}&\gls*{t1}-\gls*{t2}-\gls*{mt}&0.856$\pm$0.055&\textbf{0.867$\pm$0.052}&0.161$\pm$0.087&0.652$\pm$0.159\\
\hline
{\cellcolor{white}}&\gls*{t1}&0.932$\pm$0.006&0.931$\pm$0.012&0.066$\pm$0.012&0.854$\pm$0.014\\
                    &\gls*{t1}-\gls*{t2}&0.937$\pm$0.010&0.932$\pm$0.024&0.058$\pm$0.017&0.865$\pm$0.024\\
{\cellcolor{white}}&\gls*{t1}-\gls*{mt}&0.939$\pm$0.011&0.932$\pm$0.030&\textbf{0.053$\pm$0.017}&0.869$\pm$0.025\\
\multirow{-4}{*}{WM}&\gls*{t1}-\gls*{t2}-\gls*{mt}&\textbf{0.946$\pm$0.007}&\textbf{0.952$\pm$0.014}&0.061$\pm$0.025&\textbf{0.885$\pm$0.016}\\
\hline
\end{tabularx} 
\end{table*}
A summary of the indices evaluated over the hard segmentations is shown
  in \autoref{table:rep_overlap}.
For the \gls*{ev_fsi} \eqref{eq:si_index} counterpart, visual plots are presented
  in \autoref{fig:rep_overlap}.
The results are consistent with the conclusion drawn from the previous experiment,
  that is, a slight advantage of two-channel segmentation over classical monospectral
  and three-channel segmentations.
Specifically, the combination of  \gls*{t1} with \gls*{mt} showed better results than
  the remaining choices (despite a few outliers).
This conclusion is supported by recent work studying how \gls*{mt} can improve brain tissue
  segmentation with respect to using \gls*{t1} alone \citep{helms_improved_2009}.

\subsection{Suitability for large-scale studies}
\label{sec:ixi}
\begin{figure*}
  \includegraphics[width=1.0\linewidth]{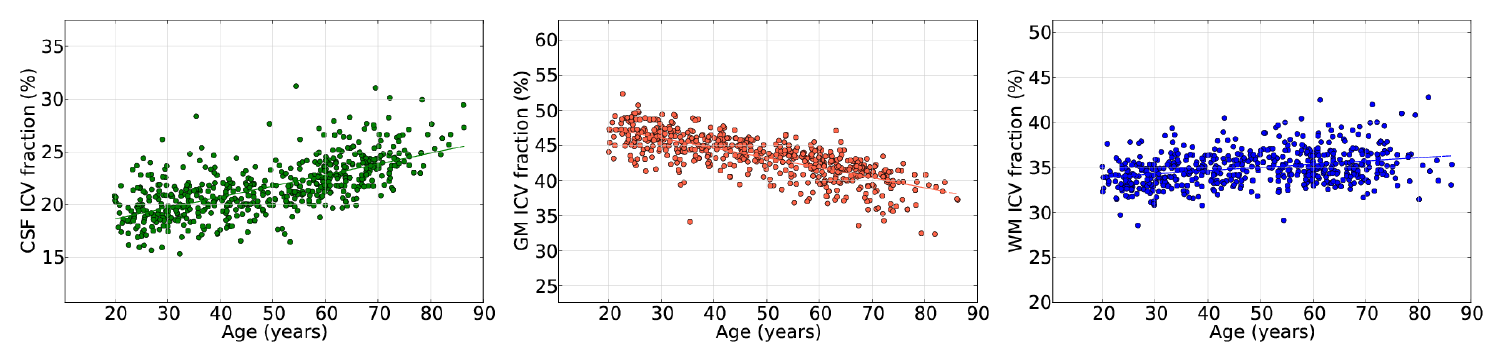}
  \caption[Volumetry study upon \emph{IXI dataset}]{Results
    for 584 of 585 subjects are presented, along with the correspondent regression lines.
    \Acrfull*{icv} fraction quantifies tissue volume with respect to the whole-brain
    volume, as defined in \autoref{sec:experimental_framework}.}
  \label{fig:ixi_regression}
\end{figure*}

\paragraph{Data}
The \emph{IXI dataset} \citep{hill_ixi_2006} is a publicly available database
  containing nearly 600 \gls*{mri} scans of healthy subjects.
The acquisition protocol for each subject includes \gls*{t1}, \gls*{t2} and \gls*{pd}
  images and some other modalities, which we did not consider in our current study.
Additional information about this database can be found in \autoref{table:data}.
From this resource, we discarded those subjects for whom information on their age was not
  available in the demographic spreadsheet distributed along with the IXI dataset.
After this reduction, a total cohort of 585 was selected for the experiment,
  which consisted of measurements of tissue volumes for all individuals to
  illustrate volume change with respect to the subjects' age
  (see \autoref{fig:ixi_regression}).

\paragraph{Image preprocessing}
We focused again on providing a standard processing pipeline, reusing part of the
  workflow defined for the previous experiment.
Firstly, we corrected for bias with the same procedure.
We accomplished the skull-stripping task on the \gls*{t1} image combining the results
  obtained with \gls*{bet} and Freesurfer (\texttt{mri\_watershed})
  for better precision.
\Gls*{t1} images were also enhanced as in the previous experiment.
Finally, intra-subject registrations of \gls*{t2} and \gls*{pd} to the reference 
  (\gls*{t1}) were performed using \gls*{flirt}.

\paragraph{Results}
We computed the \gls*{icv} fraction as defined in \autoref{sec:experimental_framework} from
  segmented data and present results in scatter plots with respect to the subjects' age
  corresponding to each dataset (see \autoref{fig:ixi_regression}).
Among the 585 subjects, segmentation failed in one case, and thus it was removed from the
  computation of the linear regression.
Our results were perfectly aligned with those published previously \citep{taki_correlations_2011,
  abe_sex_2010,mortamet_effects_2005,good_voxel-based_2001,ge_age-related_2002}.
In summary, we captured the linear increase of \gls*{csf} volume and the natural loss of \gls*{gm}
  through aging.
In addition, we perceive a more ``quadratic'' behavior of the \gls*{wm} fraction:
  slightly increasing until an age of approximately 45 years and decreasing thereafter.
As we studied relative \gls*{icv} fractions, this late decrease effect on \gls*{wm} does 
  not imply necessarily a reduction of its absolute tissue volume.
In this regard, we recall that the aim of this third study was not to show the proven 
  relationship between tissue volume and subject age.
We instead demonstrated the aptness of exploiting all of the available data with
  the multivariate approach as a useful improvement of the existing methodologies.
Therefore, we propose \gls*{mbis} as an appropriate tool for this kind of study among others.

\makeatletter{}\section{Discussion}
\label{sec:discussion}
We here present \gls*{mbis}, a segmentation tool particularly designed for
  multivariate data, and based on the Bayesian framework.
\Gls*{mbis} includes as main methodological novelties a new approach to 
  bias correction and the \gls*{markov} model optimization using
  \gls*{graph_cuts}.
After reviewing the theoretical background and implementation details,
  we reported an study evaluating the accuracy, with comparison to a widely
  used similar tool (\gls*{fast}).
Finally, we demonstrated the robustness of \gls*{mbis} on two publicly available
  multivariate databases.

\subsection{Accuracy performance}
Both visual (\autoref{fig:brainweb_hard}, \autoref{fig:brainweb_fuzzy}) and
  quantitative (\autoref{table:brainweb}) results showed the accuracy
  of the tool.
We recall that this claim was restricted to only one model from a synthetic brain
  database.
Many studies \citep{cuadra_comparison_2005,de_boer_accuracy_2010,ashburner_unified_2005,
  klauschen_evaluation_2009} that evaluated the accuracy of monospectral segmentation
  methods on \gls*{t1} \gls*{mri} have been reported.
In general, these studies used the 20 normal models from the BrainWeb 
  \citep{aubert-broche_twenty_2006}, evaluation tools \citep{shattuck_online_2009},
  or a number of manually segmented studies as ground-truth.
Nonetheless, the BrainWeb database only provides multivariate datasets for one
  single model, manual segmentation is unaffordable in multivariate data, and
  to our knowledge, there are no other evaluation resources of multivariate images.
Thus, there is an important lack of realistic ground-truth data to test
  multivariate segmentation of the brain.
Additionally, quantitative assessment of accuracy can be discredited in two ways.
First, synthetic models may not capture the unpredictable complexity of the real
  data supplied by a healthy or diseased human brain (i.e. foldings, \gls*{mri} 
  contrast properties).
Second, manual segmentation of real data taken as gold-standard
  is unaffordable, or at least, prone to inaccuracy and inconsistency
  that are intrinsic to the methodology itself and the effort-demanding nature of manual
  segmentation.
Therefore, evaluation experiments based on this scheme are illustrative
   but not definitive.
As an informal corollary, we claim that once the segmentation accuracy has been 
  assessed, it is equally or even more important to explore the challenging issue 
  of repeatability of results.
The reproducibility problem has become a main focus of interest 
  \citep{landman_effects_2007,de_boer_accuracy_2010} in every medical imaging study 
  because its absence has more negative consequences than inaccuracy itself.

Besides the segmentation results, we presented a new application of a B-spline model for
  the bias estimation.
Built upon two existing methods, we combined the original bias estimation methodology
  described previously \citep{van_leemput_automated_1999} with the bias field model proposed
  in another study \citep{tustison_n4itk:_2010}.
The B-spline model of \citeauthor{tustison_n4itk:_2010} has been proven to behave accurately
  without a heavy computational cost, and it was naturally embedded within the \gls*{e_m} 
  algorithm, as described in \autoref{sec:em}.
Visual assessment of results was documented.
Our methodology improved the bias estimated with respect to \gls*{fast} for all the channels.
Moreover, multivariate segmentation performed robustly against the bias field.
This is justified because the image channels share a unique distribution model that is used to
  estimate the bias more effectively, regardless of the modality of the channel.
Even though we observed that the bias correction can have a slightly negative impact
  on final segmentation, we concluded that the explicit modeling of the bias field is
  interesting as a multichannel bias field estimation technique itself.
Most of bias correction methodologies (e.g. \cite{tustison_n4itk:_2010}) have been well tested on
  \gls*{t1} images, but their behavior has not been studied in depth with other \gls*{mri} sequences.
In addition, they do not exploit the advantages of the underlying distribution model shared
  among pulse sequences.
The fine tuning of the bias estimation strategy included in \gls*{mbis}, and the demonstration
  of its expedience for the bias correction of multivariate images is a promising
  line for forthcoming research.

The robustness against bias field inhomogeneity exhibited by the multivariate segmentation technique
  was an illustrative example for promoting the use of multivariate approaches in neuroimaging.

\subsection{Repeatability analysis}
We tested the robustness of the presented tool against the variabilities that are intrinsic,
  mainly, to the acquisition.
We used the \emph{Kirby21 database}, consisting of 21 scan-rescan sessions on 21 healthy subjects
  with a multivariate protocol.
We attempted a number of combinations of the most suitable modalities
  (\gls*{t1}, \gls*{t2} and \gls*{mt}) for our experiment.
The results highlighted three important limitations in the experiments:
1) The quality of the image channels affects the sensitivity and robustness achieved in the 
   experiment.
2) Registration between observed variables ideally needs to be perfect.
  We briefly studied this drawback within the first experiment and present the
  impact in \autoref{fig:segmentation_pitfalls}.
  Despite this deteriorating event, the results remained satisfactory.
3) The measured increments of \acrfull*{icv} fractions strongly relied
   on the brain mask obtained with the brain extraction tool.
   Important differences in the volume of this mask
   biased the quantitative results. Less impact should be
   expected in the comparison between modality combinations.

The intention was to replicate the robustness analyses proposed elsewhere
  \citep{de_boer_accuracy_2010}.
However, one of our design considerations was to present results
  on publicly available data.
As we use a different database, the results presented in this work are not
  directly comparable to this previous study.
  
\subsection{Aptness of \gls*{mbis} in large-scale studies}
We performed a large-scale study with 585 cases, from a freely available database.
This experiment showed the expedience of \gls*{mbis} for the robust segmentation of large
  volumes of data, producing sound results.
Of note, our experiment was not intended to contribute to the field of study of brain aging,
  but we proved that \gls*{mbis} can be used for this purpose in a large-scale study with
  multivariate data (i.e. \citep{hodneland_automated_2012}).
Further work may prove that this multivariate approach is better than traditional \gls*{t1}-based 
  analysis, but in this case, we were restricted by the public availability of datasets.
As addressed in \autoref{sec:potentials_multivariate} below, this paper is intended to open
  the discussion of the potential benefits of multivariate analysis.
By selecting the appropriate MR contrasts, and developing new models for the selected multivariate
  distributions, the results of this last experiment should significantly outperform the classical
  monospectral approach.

\subsection{Potential of multivariate segmentation}
\label{sec:potentials_multivariate}
Besides the free availability of the presented tool and its evaluation,
  the most interesting result was the potential robustness suggested by
  multivariate segmentation.
There are still some challenging issues in brain tissue segmentation, for
  example the need for precise delineation of deep \gls*{gm} structures.
Some efforts have been devoted to deep brain nuclei segmentation 
  \citep{pohl_hierarchical_2007,tu_brain_2008}, but this
  application was beyond the scope of this work, given that, in general, brain
  tissue segmentation is not aimed at identifying the nuclei.
Some studies more aligned with \gls*{mbis} foundations proposed new acquisition
  sequences \citep{marques_mp2rage_2010,west_novel_2012} or the use of some 
  other existing ones \citep{helms_improved_2009} to overcome this issue.
Many of these sequences are acquired implicitly registered with other modalities,
  while some are inherently multichannel, which necessitates fully supportive
  multivariate segmentation.
Consequently, the results presented in this paper using well-established modalities could
  be improved by those obtained with the aforementioned emerging
  modalities and multivariate sequences.

The robustness issued by priors in atlas-based methods can be achieved
  with multivariate segmentation without atlases, overcoming the drawbacks of
  monospectral data-driven methods.
As mentioned in \autoref{sec:introduction}, atlas-free Bayesian segmentation methods
  can be directly applied in the clinical assessment of several global pathologies (e.g.
  atrophy, degeneration, enlarged ventricles) without modifications.
In focal conditions (e.g. tumors, multiple sclerosis, white matter lesions),
  the main requirement is the adaptation of the model used in normal subjects to the
  pathology, or including outlier rejection schemes \citep{van_leemput_automated_2001}.
In this context, \gls*{mbis} is certainly a potentially useful tool, given its
  availability and its flexibility for the necessary adaptations.

\section{Conclusion}
This work presented new and flexible segmentation software that is
  intended as the basis for a large statistical clustering
  suite for biomedical imaging.
To this end, a first version of the \gls*{mbis} tool has been made publicly 
  available, providing clinical researchers with a complete and functional
 software tool and a complementary testing framework which includes the
 presented experiments.
We also presented \gls*{mbis} to encourage multivariate analysis of data,
  as an emerging set of methodologies that should eventually
  improve the repeatability of segmentation procedures.
Complexity sources on multivariate statistical clustering are plentiful, with
  numerous alternatives, such as the mixture model selection,
  the \gls*{markov} model optimization, the bias field estimation 
  and correction, and the use of atlases.
The presented first release of \gls*{mbis} supports $n$-class segmentation using
  the \gls*{e_m} algorithm, with a novel bias modeling approach, and
  \gls*{markov} model regularization solved with \gls*{graph_cuts}
  optimization.
We evaluated the accuracy and robustness of \gls*{mbis} to demonstrate its usefulness.
Finally, we understand that \gls*{mbis} will be useful for both computer vision as
  well as clinical communities; we also hope that it will eventually encourage
  investigators to enhance further the capabilities of this publicly available 
  research tool. 

\makeatletter{}\section*{Information Sharing Statement}
\label{sec:iss}
\paragraph{Mode of Availability}
\gls*{mbis} source code is publicly available at \url{https://github.com/oesteban/MBIS}.
Data used in this work are found in the mentioned databases.
\paragraph{Requirements and specifications}
The software uses the \emph{CMake} build system (\url{http://www.cmake.org}),
  enabling compilation for all the platforms supported.
\Gls*{mbis} requires the following libraries:
  C++ and C++ Standard Library,
  \emph{Boost} Program Options, Filesystem and System Libraries \url{http://www.boost.org}
  \gls*{itk}-4.2  \url{http://www.itk.org/Wiki/ITK/Git/Download},
  maxflow-3.0.2 (only for research purposes, \url{http://pub.ist.ac.at/~vnk/software.html}), or
  maxflow-2.2.1 (GPL license, \url{http://pub.ist.ac.at/~vnk/software/maxflow-v2.21.src.tar.gz}).
The evaluation framework has the following dependencies:
  Python 2.7 (\url{http://www.python.org/}),
  nibabel (\url{http://nipy.sourceforge.net/nibabel/}),
  numpy (\url{http://numpy.scipy.org/}),
  \gls*{nipype} (\url{http://nipy.sourceforge.net/nipype/index.html}),
  matplotlib (\url{http://matplotlib.sourceforge.net/}), and
  PyPR (\url{http://pypr.sourceforge.net/}).
  
\paragraph{License}
This software is released under the \gls*{gpl} version 3  (\url{http://www.gnu.org/licenses/}). A copy of the \gls*{gpl} is distributed along with \gls*{mbis}. 

\makeatletter{}\section*{Acknowledgements}
\label{sec:acks}
The authors want to thank Tobias Kober\footnote{Advanced Clinical Imaging 
Technology, Siemens Healthcare Sector IM\&WS S, Lausanne, Switzerland} and 
Alexis Roche\footnotemark[\value{footnote}]\textsuperscript{,}\footnote{Centre 
Hospitalier Universitaire Vaudois and University of Lausanne, Switzerland} for 
their helpful discussions and suggestions. Rosie de Pietro kindly helped in the review
process. Finally, we especially thank Zaldy S. Tan for his deep revision
of the work.
 
 This study was supported by Spain's Ministry of Science and Innovation
  (projects TEC2010-21619-C04-03, TEC2011-28972-C02-02, CDTI-CENIT AMIT and
  INNPACTO PRECISION),
  Comunidad de Madrid (ARTEMIS) and
  European Regional Development Funds,
  Center for Biomedical Imaging (CIBM) of the Geneva and Lausanne Universities
  and the EPFL,
  as well as the Leenaards and Louis Jeantet Foundations,
  and the Federal Commission for Scholarships for Foreign Students of the
  Swiss Government (ESKAS 2011-12).

\FloatBarrier
\bibliographystyle{cmpb}
\bibliography{COMM_3778}

\begin{thebibliography}{84}
\providecommand{\natexlab}[1]{#1}
\providecommand{\url}[1]{\texttt{#1}}
\expandafter\ifx\csname urlstyle\endcsname\relax
  \providecommand{\doi}[1]{doi: #1}\else
  \providecommand{\doi}{doi: \begingroup \urlstyle{rm}\Url}\fi

\bibitem[Abe et~al.(2010)Abe, Yamasue, Yamada, Masutani, Kabasawa, Sasaki,
  Takei, Suga, Kasai, and Aoki]{abe_sex_2010}
O.~Abe, H.~Yamasue, H.~Yamada, Y.~Masutani, H.~Kabasawa, H.~Sasaki, K.~Takei,
  M.~Suga, K.~Kasai, and S.~Aoki.
\newblock Sex dimorphism in gray/white matter volume and diffusion tensor
  during normal aging.
\newblock \emph{{NMR} Biomed}, 23\penalty0 (5):\penalty0 446--458, 2010.

\bibitem[Ahmed et~al.(2002)Ahmed, Yamany, Mohamed, Farag, and
  Moriarty]{ahmed_modified_2002}
M.~Ahmed, S.~Yamany, N.~Mohamed, A.~Farag, and T.~Moriarty.
\newblock A modified fuzzy c-means algorithm for bias field estimation and
  segmentation of {MRI} data.
\newblock \emph{{IEEE} Trans Med Imag}, 21\penalty0 (3):\penalty0 193--199,
  2002.
\newblock \doi{10.1109/42.996338}.

\bibitem[Altman and Bland(1983)]{altman_measurement_1983}
D.~G. Altman and J.~M. Bland.
\newblock Measurement in medicine: The analysis of method comparison studies.
\newblock \emph{J R Stat Soc (S-D, Statist)}, 32\penalty0 (3):\penalty0
  307--317, 1983.
\newblock \doi{10.2307/2987937}.

\bibitem[Anbeek et~al.(2004)Anbeek, Vincken, van Osch, Bisschops, and van~der
  Grond]{anbeek_probabilistic_2004}
P.~Anbeek, K.~L. Vincken, M.~J.~P. van Osch, R.~H.~C. Bisschops, and J.~van~der
  Grond.
\newblock Probabilistic segmentation of white matter lesions in {MR} imaging.
\newblock \emph{{NeuroImage}}, 21\penalty0 (3):\penalty0 1037--1044, 2004.
\newblock \doi{10.1016/j.neuroimage.2003.10.012}.

\bibitem[Ashburner and Friston(2001)]{ashburner_why_2001}
J.~Ashburner and K.~J. Friston.
\newblock Why voxel-based morphometry should be used.
\newblock \emph{{NeuroImage}}, 14\penalty0 (6):\penalty0 1238--1243, 2001.
\newblock \doi{10.1006/nimg.2001.0961}.

\bibitem[Ashburner and Friston(2005)]{ashburner_unified_2005}
J.~Ashburner and K.~J. Friston.
\newblock Unified segmentation.
\newblock \emph{{NeuroImage}}, 26\penalty0 (3):\penalty0 839--851, 2005.
\newblock \doi{10.1016/j.neuroimage.2005.02.018}.

\bibitem[Aubert-Broche et~al.(2006)Aubert-Broche, Griffin, Pike, Evans, and
  Collins]{aubert-broche_twenty_2006}
B.~Aubert-Broche, M.~Griffin, G.~B. Pike, A.~C. Evans, and D.~L. Collins.
\newblock Twenty new digital brain phantoms for creation of validation image
  data bases.
\newblock \emph{{IEEE} Trans Med Imag}, 25\penalty0 (11):\penalty0 1410--1416,
  2006.
\newblock \doi{10.1109/TMI.2006.883453}.

\bibitem[Avants et~al.(2013)Avants, Song, Duda, Johnson, and
  Tustison]{avants_ants:_2013}
B.~Avants, G.~Song, J.~Duda, H.~Johnson, and N.~Tustison.
\newblock {ANTs:} advanced normalization tools, 2013.
\newblock URL \url{http://www.picsl.upenn.edu/ANTS/}.
\newblock Last accessed May 7th, 2013.

\bibitem[Avants et~al.(2011)Avants, Tustison, Wu, Cook, and
  Gee]{avants_open_2011}
B.~B. Avants, N.~J. Tustison, J.~Wu, P.~A. Cook, and J.~C. Gee.
\newblock An open source multivariate framework for n-tissue segmentation with
  evaluation on public data.
\newblock \emph{Neuroinformatics}, 9\penalty0 (4):\penalty0 381--400, 2011.
\newblock \doi{10.1007/s12021-011-9109-y}.

\bibitem[Bishop(2009)]{bishop_pattern_2009}
C.~M. Bishop.
\newblock \emph{Pattern recognition and machine learning}.
\newblock Springer, New York, {NY}, 2009.

\bibitem[Bookstein(2001)]{bookstein_voxel-based_2001}
F.~L. Bookstein.
\newblock {"Voxel-Based} morphometry" should not be used with imperfectly
  registered images.
\newblock \emph{{NeuroImage}}, 14\penalty0 (6):\penalty0 1454--1462, 2001.
\newblock \doi{10.1006/nimg.2001.0770}.

\bibitem[Boykov and Kolmogorov(2004)]{boykov_experimental_2004}
Y.~Boykov and V.~Kolmogorov.
\newblock An experimental comparison of min-cut/max-flow algorithms for energy
  minimization in vision.
\newblock \emph{{IEEE} Trans Pattern Anal Mach Intell}, 26\penalty0
  (9):\penalty0 1124--1137, 2004.
\newblock \doi{10.1109/TPAMI.2004.60}.

\bibitem[Boykov et~al.(2001)Boykov, Veksler, and Zabih]{boykov_fast_2001}
Y.~Boykov, O.~Veksler, and R.~Zabih.
\newblock Fast approximate energy minimization via graph cuts.
\newblock \emph{{IEEE} Trans Pattern Anal Mach Intell}, 23\penalty0
  (11):\penalty0 1222--1239, 2001.
\newblock \doi{10.1109/34.969114}.

\bibitem[Bromiley and Thacker(2008)]{bromiley_multi-dimensional_2008}
P.~A. Bromiley and N.~A. Thacker.
\newblock Multi-dimensional medical image segmentation with partial volume and
  gradient modelling.
\newblock \emph{Annals {BMVA}}, 2\penalty0 (1-22):\penalty0 2--4, 2008.

\bibitem[Cardoso et~al.(2012)Cardoso, Clarkson, Modat, and
  Ourselin]{cardoso_niftyseg:_2012}
M.~J. Cardoso, M.~J. Clarkson, M.~Modat, and S.~Ourselin.
\newblock {NiftySeg:} open-source software for medical image segmentation,
  label fusion and cortical thickness estimation.
\newblock In \emph{{IEEE} International Symposium on Biomedical Imaging},
  Barcelona, Spain, 2012.

\bibitem[Cocosco et~al.(1997)Cocosco, Kollokian, Kwan, and
  Evans]{cocosco_brainweb:_1997}
C.~Cocosco, V.~Kollokian, R.~Kwan, and A.~Evans.
\newblock {BrainWeb:} online interface to a {3D} {MRI} simulated brain
  database.
\newblock In \emph{Anual Meeting of the Org Hum Brain Mapp}, vol. 5(4), Part 2,
  S425, Copenhangen, Denmark, 1997. {NeuroImage}.
\newblock \doi{10.1016/S1053-8119(97)80018-3}.

\bibitem[Collins et~al.(2001)Collins, Montagnat, Zijdenbos, Evans, and
  Arnold]{collins_automated_2001}
D.~L. Collins, J.~Montagnat, A.~P. Zijdenbos, A.~C. Evans, and D.~L. Arnold.
\newblock Automated estimation of brain volume in multiple sclerosis with
  {BICCR}.
\newblock In M.~F. Insana and R.~M. Leahy, editors, \emph{Information
  Processing in Medical Imaging ({IPMI)}}, vol. 2082 of \emph{{LNCS}}, pp.
  141--147, Davis, {CA}, {USA}, 2001. Springer Berlin Heidelberg.
\newblock \doi{10.1007/3-540-45729-1_12}.

\bibitem[Crum et~al.(2003)Crum, Griffin, Hill, and Hawkes]{crum_zen_2003}
W.~Crum, L.~Griffin, D.~Hill, and D.~Hawkes.
\newblock Zen and the art of medical image registration: correspondence,
  homology, and quality.
\newblock \emph{{NeuroImage}}, 20\penalty0 (3):\penalty0 1425--1437, 2003.
\newblock \doi{10.1016/j.neuroimage.2003.07.014}.

\bibitem[Crum et~al.(2006)Crum, Camara, and Hill]{crum_generalized_2006}
W.~Crum, O.~Camara, and D.~Hill.
\newblock Generalized overlap measures for evaluation and validation in medical
  image analysis.
\newblock \emph{{IEEE} Trans Med Imag}, 25\penalty0 (11):\penalty0 1451--1461,
  2006.
\newblock \doi{10.1109/TMI.2006.880587}.

\bibitem[Cuadra et~al.(2005)Cuadra, Cammoun, Butz, Cuisenaire, and
  Thiran]{cuadra_comparison_2005}
M.~B. Cuadra, L.~Cammoun, T.~Butz, O.~Cuisenaire, and J.~P. Thiran.
\newblock Comparison and validation of tissue modelization and statistical
  classification methods in {T1} weighted {MR} brain images.
\newblock \emph{{IEEE} Trans Med Imag}, 24\penalty0 (12):\penalty0 1548--1565,
  2005.
\newblock \doi{10.1109/TMI.2005.857652}.

\bibitem[Dang et~al.(2013)Dang, Modi, Roberts, Chan, and
  Mitchell]{dang_validation_2013}
M.~Dang, J.~Modi, M.~Roberts, C.~Chan, and J.~R. Mitchell.
\newblock Validation study of a fast, accurate, and precise brain tumor volume
  measurement.
\newblock \emph{Comp Meth Prog Bio}, 111\penalty0 (2):\penalty0 480--487, 2013.
\newblock \doi{10.1016/j.cmpb.2013.04.011}.

\bibitem[Davatzikos(2004)]{davatzikos_why_2004}
C.~Davatzikos.
\newblock Why voxel-based morphometric analysis should be used with great
  caution when characterizing group differences.
\newblock \emph{{NeuroImage}}, 23\penalty0 (1):\penalty0 17--20, 2004.
\newblock \doi{10.1016/j.neuroimage.2004.05.010}.

\bibitem[De~Boer et~al.(2010)De~Boer, Vrooman, Ikram, Vernooij, Breteler,
  Van~der Lugt, and Niessen]{de_boer_accuracy_2010}
R.~De~Boer, H.~A. Vrooman, M.~A. Ikram, M.~W. Vernooij, M.~Breteler, A.~Van~der
  Lugt, and W.~J. Niessen.
\newblock Accuracy and reproducibility study of automatic {MRI} brain tissue
  segmentation methods.
\newblock \emph{{NeuroImage}}, 51\penalty0 (3):\penalty0 1047--1056, 2010.
\newblock \doi{10.1016/j.neuroimage.2010.03.012}.

\bibitem[Delibasis et~al.(2013)Delibasis, Kechriniotis, and
  Maglogiannis]{delibasis_novel_2013}
K.~K. Delibasis, A.~Kechriniotis, and I.~Maglogiannis.
\newblock A novel tool for segmenting {3D} medical images based on generalized
  cylinders and active surfaces.
\newblock \emph{Comp Meth Prog Bio}, 111\penalty0 (1):\penalty0 148--165, 2013.
\newblock \doi{10.1016/j.cmpb.2013.03.009}.

\bibitem[Duch{\'e} et~al.(2012)Duch{\'e}, Acosta, Gambarota, Merlet, Salvado,
  and Saint-Jalmes]{duche_bi-exponential_2012}
Q.~Duch{\'e}, O.~Acosta, G.~Gambarota, I.~Merlet, O.~Salvado, and
  H.~Saint-Jalmes.
\newblock Bi-exponential magnetic resonance signal model for partial volume
  computation.
\newblock In N.~Ayache, H.~Delingette, P.~Golland, and K.~Mori, editors,
  \emph{Medical Image Computing and Computer-Assisted Intervention ({MICCAI)}},
  vol. 7510 of \emph{{LNCS}}, pp. 231--238, Nice, France, 2012. Springer Berlin
  Heidelberg.
\newblock \doi{10.1007/978-3-642-33415-3_29}.

\bibitem[Fischl(2012)]{fischl_freesurfer_2012}
B.~Fischl.
\newblock {FreeSurfer}.
\newblock \emph{{NeuroImage}}, 62\penalty0 (2):\penalty0 774--781, 2012.
\newblock \doi{10.1016/j.neuroimage.2012.01.021}.

\bibitem[Fischl and Dale(2000)]{fischl_measuring_2000}
B.~Fischl and A.~M. Dale.
\newblock Measuring the thickness of the human cerebral cortex from magnetic
  resonance images.
\newblock \emph{{PNAS}}, 97\penalty0 (20):\penalty0 11050--11055, 2000.
\newblock \doi{10.1073/pnas.200033797}.

\bibitem[Fischl et~al.(2002)Fischl, Salat, Busa, Albert, Dieterich, Haselgrove,
  van~der Kouwe, Killiany, Kennedy, and Klaveness]{fischl_whole_2002}
B.~Fischl, D.~H. Salat, E.~Busa, M.~Albert, M.~Dieterich, C.~Haselgrove,
  A.~van~der Kouwe, R.~Killiany, D.~Kennedy, and S.~Klaveness.
\newblock Whole brain segmentation: automated labeling of neuroanatomical
  structures in the human brain.
\newblock \emph{Neuron}, 33\penalty0 (3):\penalty0 341--355, 2002.
\newblock \doi{10.1016/S0896-6273(02)00569-X}.

\bibitem[Ge et~al.(2002)Ge, Grossman, Babb, Rabin, Mannon, and
  Kolson]{ge_age-related_2002}
Y.~Ge, R.~I. Grossman, J.~S. Babb, M.~L. Rabin, L.~J. Mannon, and D.~L. Kolson.
\newblock Age-related total gray matter and white matter changes in normal
  adult brain. {Part} {I}: volumetric {MR} imaging analysis.
\newblock \emph{{AJNR} Am J Neuroradiol}, 23\penalty0 (8):\penalty0 1327--1333,
  2002.

\bibitem[Geman and Geman(1984)]{geman_stochastic_1984}
S.~Geman and D.~Geman.
\newblock Stochastic relaxation, {Gibbs} distributions, and the bayesian
  restoration of images.
\newblock \emph{{IEEE} Trans Pattern Anal Mach Intell}, 6\penalty0
  (6):\penalty0 721--741, 1984.
\newblock \doi{10.1109/TPAMI.1984.4767596}.

\bibitem[Good et~al.(2001)Good, Johnsrude, Ashburner, Henson, Friston, and
  Frackowiak]{good_voxel-based_2001}
C.~D. Good, I.~S. Johnsrude, J.~Ashburner, R.~N. Henson, K.~J. Friston, and
  R.~S. Frackowiak.
\newblock A voxel-based morphometric study of ageing in 465 normal adult human
  brains.
\newblock \emph{{NeuroImage}}, 14\penalty0 (1):\penalty0 21--36, 2001.
\newblock \doi{10.1006/nimg.2001.0786}.

\bibitem[Gorgolewski et~al.(2011)Gorgolewski, Burns, Madison, Clark, Halchenko,
  Waskom, and Ghosh]{gorgolewski_nipype:_2011}
K.~Gorgolewski, C.~D. Burns, C.~Madison, D.~Clark, Y.~O. Halchenko, M.~L.
  Waskom, and S.~S. Ghosh.
\newblock Nipype: a flexible, lightweight and extensible neuroimaging data
  processing framework in python.
\newblock \emph{Front Neuroinform}, 5:\penalty0 13, 2011.
\newblock \doi{10.3389/fninf.2011.00013}.

\bibitem[Gorthi et~al.(2011)Gorthi, Duay, Bresson, Bach~Cuadra,
  S{\'a}nchez~Castro, Pollo, Allal, and Thiran]{gorthi_active_2011}
S.~Gorthi, V.~Duay, X.~Bresson, M.~Bach~Cuadra, F.~J. S{\'a}nchez~Castro,
  C.~Pollo, A.~S. Allal, and J.-P. Thiran.
\newblock Active deformation fields: dense deformation field estimation for
  atlas-based segmentation using the active contour framework.
\newblock \emph{Med Image Anal}, 15\penalty0 (6):\penalty0 787{\textendash}800,
  2011.
\newblock \doi{10.1016/j.media.2011.05.008}.

\bibitem[Hazlett et~al.(2006)Hazlett, Poe, Gerig, Smith, and
  Piven]{hazlett_cortical_2006}
H.~C. Hazlett, M.~D. Poe, G.~Gerig, R.~G. Smith, and J.~Piven.
\newblock Cortical gray and white brain tissue volume in adolescents and adults
  with autism.
\newblock \emph{Biol Psychiatry}, 59\penalty0 (1):\penalty0 1--6, 2006.
\newblock \doi{10.1016/j.biopsych.2005.06.015}.

\bibitem[He et~al.(2008)He, Datta, Sajja, and Narayana]{he_generalized_2008}
R.~He, S.~Datta, B.~R. Sajja, and P.~A. Narayana.
\newblock Generalized fuzzy clustering for segmentation of multi-spectral
  magnetic resonance images.
\newblock \emph{Comput Med Imag Grap}, 32\penalty0 (5):\penalty0 353--366,
  2008.
\newblock \doi{10.1016/j.compmedimag.2008.02.002}.

\bibitem[Helms et~al.(2009)Helms, Draganski, Frackowiak, Ashburner, and
  Weiskopf]{helms_improved_2009}
G.~Helms, B.~Draganski, R.~Frackowiak, J.~Ashburner, and N.~Weiskopf.
\newblock Improved segmentation of deep brain grey matter structures using
  magnetization transfer ({MT)} parameter maps.
\newblock \emph{{NeuroImage}}, 47\penalty0 (1):\penalty0 194--198, 2009.
\newblock \doi{10.1016/j.neuroimage.2009.03.053}.

\bibitem[Hill et~al.(2006)Hill, Williams, Hawkes, and Smith]{hill_ixi_2006}
D.~Hill, S.~Williams, D.~Hawkes, and S.~Smith.
\newblock {IXI} dataset - {Information} {eXtraction} from {Images} project,
  2006.
\newblock URL \url{http://www.brain-development.org/}.
\newblock Last accessed May 7th, 2013.

\bibitem[Hodneland et~al.(2012)Hodneland, Ystad, Haasz, Munthe-Kaas, and
  Lundervold]{hodneland_automated_2012}
E.~Hodneland, M.~Ystad, J.~Haasz, A.~Munthe-Kaas, and A.~Lundervold.
\newblock Automated approaches for analysis of multimodal {MRI} acquisitions in
  a study of cognitive aging.
\newblock \emph{Comp Meth Prog Bio}, 106\penalty0 (3):\penalty0 328--341, 2012.
\newblock \doi{10.1016/j.cmpb.2011.03.010}.

\bibitem[Ibanez et~al.(2006)Ibanez, Avila, and Aylward]{ibanez_open_2006}
L.~Ibanez, R.~Avila, and S.~Aylward.
\newblock Open source and open science: how it is changing the medical imaging
  community.
\newblock In \emph{{IEEE} International Symposium on Biomedical Imaging
  ({ISBI)}}, pp. 690--693, Arlington, {VA}, {USA}, 2006.
\newblock \doi{10.1109/ISBI.2006.1625010}.

\bibitem[Jenkinson et~al.(2002)Jenkinson, Bannister, Brady, and
  Smith]{jenkinson_improved_2002}
M.~Jenkinson, P.~Bannister, M.~Brady, and S.~Smith.
\newblock Improved optimization for the robust and accurate linear registration
  and motion correction of brain images.
\newblock \emph{{NeuroImage}}, 17\penalty0 (2):\penalty0 825--841, 2002.
\newblock \doi{10.1006/nimg.2002.1132}.

\bibitem[Ji et~al.(2012)Ji, Sun, Xia, Chen, Xia, and Feng]{ji_generalized_2012}
Z.~Ji, Q.~Sun, Y.~Xia, Q.~Chen, D.~Xia, and D.~Feng.
\newblock Generalized rough fuzzy c-means algorithm for brain {MR} image
  segmentation.
\newblock \emph{Comp Meth Prog Bio}, 108\penalty0 (2):\penalty0 644--655, 2012.
\newblock \doi{10.1016/j.cmpb.2011.10.010}.

\bibitem[Jones et~al.(2000)Jones, Buchbinder, and
  Aharon]{jones_three-dimensional_2000}
S.~E. Jones, B.~R. Buchbinder, and I.~Aharon.
\newblock Three-dimensional mapping of cortical thickness using {Laplace}'s
  equation.
\newblock \emph{Hum Brain Mapp}, 11\penalty0 (1):\penalty0 12--32, 2000.
\newblock \doi{10.1002/1097-0193(200009)11:1<12::AID-HBM20>3.0.CO;2-K}.

\bibitem[Kapur et~al.(1996)Kapur, Grimson, Wells~{III}, and
  Kikinis]{kapur_segmentation_1996}
T.~Kapur, W.~L. Grimson, W.~M. Wells~{III}, and R.~Kikinis.
\newblock Segmentation of brain tissue from magnetic resonance images.
\newblock \emph{Med Image Anal}, 1\penalty0 (2):\penalty0 109--127, 1996.
\newblock \doi{10.1016/S1361-8415(96)80008-9}.

\bibitem[Kikinis et~al.(1996)Kikinis, Shenton, Iosifescu, {McCarley},
  Saiviroonporn, Hokama, Robatino, Metcalf, Wible, Portas, Donnino, and
  Jolesz]{kikinis_digital_1996}
R.~Kikinis, M.~Shenton, D.~Iosifescu, R.~{McCarley}, P.~Saiviroonporn,
  H.~Hokama, A.~Robatino, D.~Metcalf, C.~Wible, C.~Portas, R.~Donnino, and
  F.~Jolesz.
\newblock A digital brain atlas for surgical planning, model-driven
  segmentation, and teaching.
\newblock \emph{{IEEE} Trans Vis Comput Graph}, 2\penalty0 (3):\penalty0
  232--241, 1996.
\newblock \doi{10.1109/2945.537306}.

\bibitem[Klauschen et~al.(2009)Klauschen, Goldman, Barra, Meyer-Lindenberg, and
  Lundervold]{klauschen_evaluation_2009}
F.~Klauschen, A.~Goldman, V.~Barra, A.~Meyer-Lindenberg, and A.~Lundervold.
\newblock Evaluation of automated brain {MR} image segmentation and volumetry
  methods.
\newblock \emph{Hum Brain Mapp}, 30\penalty0 (4):\penalty0 1310--1327, 2009.
\newblock \doi{10.1002/hbm.20599}.

\bibitem[Kolmogorov and Zabin(2004)]{kolmogorov_what_2004}
V.~Kolmogorov and R.~Zabin.
\newblock What energy functions can be minimized via graph cuts?
\newblock \emph{{IEEE} Trans Pattern Anal Mach Intell}, 26\penalty0
  (2):\penalty0 147--159, 2004.
\newblock \doi{10.1109/TPAMI.2004.1262177}.

\bibitem[Landman et~al.(2007)Landman, Farrell, Jones, Smith, Prince, and
  Mori]{landman_effects_2007}
B.~A. Landman, J.~A.~D. Farrell, C.~K. Jones, S.~A. Smith, J.~L. Prince, and
  S.~Mori.
\newblock Effects of diffusion weighting schemes on the reproducibility of
  {DTI-derived} fractional anisotropy, mean diffusivity, and principal
  eigenvector measurements at {1.5T}.
\newblock \emph{{NeuroImage}}, 36\penalty0 (4):\penalty0 1123--1138, 2007.
\newblock \doi{10.1016/j.neuroimage.2007.02.056}.

\bibitem[Landman et~al.(2011)Landman, Huang, Gifford, Vikram, Lim, Farrell,
  Bogovic, Hua, Chen, Jarso, Smith, Joel, Mori, Pekar, Barker, Prince, and van
  Zijl]{landman_multi-parametric_2011}
B.~A. Landman, A.~J. Huang, A.~Gifford, D.~S. Vikram, I.~A.~L. Lim, J.~A.
  Farrell, J.~A. Bogovic, J.~Hua, M.~Chen, S.~Jarso, S.~A. Smith, S.~Joel,
  S.~Mori, J.~J. Pekar, P.~B. Barker, J.~L. Prince, and P.~C. van Zijl.
\newblock Multi-parametric neuroimaging reproducibility: A {3T} resource study.
\newblock \emph{{NeuroImage}}, 54\penalty0 (4):\penalty0 2854--2866, 2011.
\newblock \doi{10.1016/j.neuroimage.2010.11.047}.

\bibitem[Li(2009)]{li_markov_2009}
S.~Z. Li.
\newblock \emph{Markov Random Field Modeling in Image Analysis}.
\newblock Advances in Pattern Recognition. Springer London, 2009.

\bibitem[Liang and Wang(2009)]{liang_em_2009}
Z.~Liang and S.~Wang.
\newblock An {EM} approach to {MAP} solution of segmenting tissue mixtures: A
  numerical analysis.
\newblock \emph{{IEEE} Trans Med Imag}, 28\penalty0 (2):\penalty0 297--310,
  2009.
\newblock \doi{10.1109/TMI.2008.2004670}.

\bibitem[Liew and Yan(2006)]{liew_current_2006}
A.~W.-C. Liew and H.~Yan.
\newblock Current methods in the automatic tissue segmentation of {3D} magnetic
  resonance brain images.
\newblock \emph{Curr Med Imag Rev}, 2\penalty0 (1):\penalty0 91--103, 2006.

\bibitem[{MacDonald} et~al.(2000){MacDonald}, Kabani, Avis, and
  Evans]{macdonald_automated_2000}
D.~{MacDonald}, N.~Kabani, D.~Avis, and A.~C. Evans.
\newblock Automated {3D} extraction of inner and outer surfaces of cerebral
  cortex from {MRI}.
\newblock \emph{{NeuroImage}}, 12\penalty0 (3):\penalty0 340--356, 2000.
\newblock \doi{10.1006/nimg.1999.0534}.

\bibitem[Marques et~al.(2010)Marques, Kober, Krueger, van~der Zwaag, Van~de
  Moortele, and Gruetter]{marques_mp2rage_2010}
J.~P. Marques, T.~Kober, G.~Krueger, W.~van~der Zwaag, P.-F. Van~de Moortele,
  and R.~Gruetter.
\newblock {MP2RAGE}, a self bias-field corrected sequence for improved
  segmentation and {{T1}}-mapping at high field.
\newblock \emph{{NeuroImage}}, 49\penalty0 (2):\penalty0 1271--1281, 2010.
\newblock \doi{10.1016/j.neuroimage.2009.10.002}.

\bibitem[Menze et~al.(2014)Menze, Jakab, Bauer, Kalpathy-Cramer, Farahani,
  Kirby, Burren, Porz, Slotboom, Wiest, Lanczi, Gerstner, Weber, Arbel, Avants,
  Ayache, Buendia, Collins, Cordier, Corso, Criminisi, Das, Delingette,
  Demiralp, Durst, Dojat, Doyle, Festa, Forbes, Geremia, Glocker, Golland, Guo,
  Hamamci, Iftekharuddin, Jena, John, Konukoglu, Lashkari, Mariz, Meier,
  Pereira, Precup, Price, Riklin-Raviv, Reza, Ryan, Schwartz, Shin, Shotton,
  Silva, Sousa, Subbanna, Szekely, Taylor, Thomas, Tustison, Unal, Vasseur,
  Wintermark, Ye, Zhao, Zhao, Zikic, Prastawa, Reyes, and
  Leemput]{menze_multimodal_2014}
B.~Menze, A.~Jakab, S.~Bauer, J.~Kalpathy-Cramer, K.~Farahani, J.~Kirby,
  Y.~Burren, N.~Porz, J.~Slotboom, R.~Wiest, L.~Lanczi, E.~Gerstner, M.-A.
  Weber, T.~Arbel, B.~Avants, N.~Ayache, P.~Buendia, L.~Collins, N.~Cordier,
  J.~Corso, A.~Criminisi, T.~Das, H.~Delingette, C.~Demiralp, C.~Durst,
  M.~Dojat, S.~Doyle, J.~Festa, F.~Forbes, E.~Geremia, B.~Glocker, P.~Golland,
  X.~Guo, A.~Hamamci, K.~Iftekharuddin, R.~Jena, N.~John, E.~Konukoglu,
  D.~Lashkari, J.~A. Mariz, R.~Meier, S.~Pereira, D.~Precup, S.~J. Price,
  T.~Riklin-Raviv, S.~Reza, M.~Ryan, L.~Schwartz, H.-C. Shin, J.~Shotton,
  C.~Silva, N.~Sousa, N.~Subbanna, G.~Szekely, T.~Taylor, O.~Thomas,
  N.~Tustison, G.~Unal, F.~Vasseur, M.~Wintermark, D.~H. Ye, L.~Zhao, B.~Zhao,
  D.~Zikic, M.~Prastawa, M.~Reyes, and K.~V. Leemput.
\newblock The multimodal brain tumor image segmentation benchmark ({BRATS)}.
\newblock 2014.
\newblock URL \url{http://hal.inria.fr/hal-00935640}.

\bibitem[Mortamet et~al.(2005)Mortamet, Zeng, Gerig, Prastawa, and
  Bullitt]{mortamet_effects_2005}
B.~Mortamet, D.~Zeng, G.~Gerig, M.~Prastawa, and E.~Bullitt.
\newblock Effects of healthy aging measured by intracranial compartment volumes
  using a designed {MR} brain database.
\newblock In \emph{Medical Image Computing and Computer-Assisted Intervention
  ({MICCAI)}}, vol. 3749 of \emph{{LNCS}}, pp. 383--391, Palm Springs, {CA},
  {USA}, 2005. Springer Berlin Heidelberg.
\newblock \doi{10.1007/11566465_48}.

\bibitem[Noe and Gee(2001)]{noe_partial_2001}
A.~Noe and J.~C. Gee.
\newblock Partial volume segmentation of cerebral {MRI} scans with mixture
  model clustering.
\newblock In M.~F. Insana and R.~M. Leahy, editors, \emph{Information
  Processing in Medical Imaging ({IPMI)}}, vol. 2082 of \emph{{LNCS}}, pp.
  423--430, Davis, {CA}, {USA}, 2001. Springer Berlin Heidelberg.
\newblock \doi{10.1007/3-540-45729-1_44}.

\bibitem[Paus et~al.(1999)Paus, Zijdenbos, Worsley, Collins, Blumenthal, Giedd,
  Rapoport, and Evans]{paus_structural_1999}
T.~Paus, A.~Zijdenbos, K.~Worsley, D.~L. Collins, J.~Blumenthal, J.~N. Giedd,
  J.~L. Rapoport, and A.~C. Evans.
\newblock Structural maturation of neural pathways in children and adolescents:
  In vivo study.
\newblock \emph{Science}, 283\penalty0 (5409):\penalty0 1908--1911, 1999.
\newblock \doi{10.1126/science.283.5409.1908}.

\bibitem[Pohl et~al.(2007)Pohl, Bouix, Nakamura, Rohlfing, {McCarley}, Kikinis,
  Grimson, Shenton, and Wells]{pohl_hierarchical_2007}
K.~M. Pohl, S.~Bouix, M.~Nakamura, T.~Rohlfing, R.~W. {McCarley}, R.~Kikinis,
  W.~E.~L. Grimson, M.~E. Shenton, and W.~M. Wells.
\newblock A hierarchical algorithm for {MR} brain image parcellation.
\newblock \emph{{IEEE} Trans Med Imag}, 26\penalty0 (9):\penalty0 1201--1212,
  2007.
\newblock \doi{10.1109/TMI.2007.901433}.

\bibitem[Prastawa et~al.(2003)Prastawa, Bullitt, Moon, Van~Leemput, and
  Gerig]{prastawa_automatic_2003}
M.~Prastawa, E.~Bullitt, N.~Moon, K.~Van~Leemput, and G.~Gerig.
\newblock Automatic brain tumor segmentation by subject specific modification
  of atlas priors.
\newblock \emph{Acad Radiol}, 10\penalty0 (12):\penalty0 1341--1348, 2003.
\newblock \doi{10.1016/S1076-6332(03)00506-3}.

\bibitem[Roche et~al.(2011)Roche, Ribes, Bach-Cuadra, and
  Kr{\"u}ger]{roche_convergence_2011}
A.~Roche, D.~Ribes, M.~Bach-Cuadra, and G.~Kr{\"u}ger.
\newblock On the convergence of {EM-like} algorithms for image segmentation
  using markov random fields.
\newblock \emph{Med Image Anal}, 15\penalty0 (6):\penalty0 830--839, 2011.
\newblock \doi{10.1016/j.media.2011.05.002}.

\bibitem[Roura et~al.(2012)Roura, Oliver, Cabezas, Vilanova, Rovira,
  Rami{\'o}-Torrent{\`a}, and Llad{\'o}]{roura_marga:_2012}
E.~Roura, A.~Oliver, M.~Cabezas, J.~C. Vilanova, {\`A}.~Rovira,
  L.~Rami{\'o}-Torrent{\`a}, and X.~Llad{\'o}.
\newblock {MARGA:} multispectral adaptive region growing algorithm for brain
  extraction on axial {MRI}.
\newblock \emph{Comp Meth Prog Bio}, 113\penalty0 (2):\penalty0 655--673, 2012.
\newblock \doi{10.1016/j.cmpb.2013.11.015}.

\bibitem[Santago and Gage(1995)]{santago_statistical_1995}
P.~Santago and H.~Gage.
\newblock Statistical models of partial volume effect.
\newblock \emph{{IEEE} Trans Image Process}, 4\penalty0 (11):\penalty0 1531
  --1540, 1995.
\newblock \doi{10.1109/83.469934}.

\bibitem[Shattuck et~al.(2009)Shattuck, Prasad, Mirza, Narr, and
  Toga]{shattuck_online_2009}
D.~W. Shattuck, G.~Prasad, M.~Mirza, K.~L. Narr, and A.~W. Toga.
\newblock Online resource for validation of brain segmentation methods.
\newblock \emph{{NeuroImage}}, 45\penalty0 (2):\penalty0 431--439, 2009.
\newblock \doi{10.1016/j.neuroimage.2008.10.066}.

\bibitem[Smith(2002)]{smith_fast_2002}
S.~M. Smith.
\newblock Fast robust automated brain extraction.
\newblock \emph{Hum Brain Mapp}, 17\penalty0 (3):\penalty0 143--155, 2002.
\newblock \doi{10.1002/hbm.10062}.

\bibitem[Suri(2000)]{suri_leaking_2000}
J.~Suri.
\newblock Leaking prevention in fast level sets using fuzzy models: an
  application in {MR} brain.
\newblock In \emph{{IEEE} {EMBS} International Conference on Information
  Technology Applications in Biomedicine}, pp. 220--225, Arlington, {VA},
  {USA}, 2000.
\newblock \doi{10.1109/ITAB.2000.892390}.

\bibitem[Taki et~al.(2011)Taki, Thyreau, Kinomura, Sato, Goto, Kawashima, and
  Fukuda]{taki_correlations_2011}
Y.~Taki, B.~Thyreau, S.~Kinomura, K.~Sato, R.~Goto, R.~Kawashima, and
  H.~Fukuda.
\newblock Correlations among brain gray matter volumes, age, gender, and
  hemisphere in healthy individuals.
\newblock \emph{{PLoS} {ONE}}, 6\penalty0 (7):\penalty0 e22734, 2011.
\newblock \doi{10.1371/journal.pone.0022734}.

\bibitem[Tanabe et~al.(1997)Tanabe, Amend, Schuff, {DiSclafani}, Ezekiel,
  Norman, Fein, and Weiner]{tanabe_tissue_1997}
J.~L. Tanabe, D.~Amend, N.~Schuff, V.~{DiSclafani}, F.~Ezekiel, D.~Norman,
  G.~Fein, and M.~W. Weiner.
\newblock Tissue segmentation of the brain in {Alzheimer} disease.
\newblock \emph{{AJNR} Am J Neuroradiol}, 18\penalty0 (1):\penalty0 115--123,
  1997.

\bibitem[Tohka et~al.(2004)Tohka, Zijdenbos, and Evans]{tohka_fast_2004}
J.~Tohka, A.~Zijdenbos, and A.~Evans.
\newblock Fast and robust parameter estimation for statistical partial volume
  models in brain {MRI}.
\newblock \emph{{NeuroImage}}, 23\penalty0 (1):\penalty0 84--97, 2004.
\newblock \doi{10.1016/j.neuroimage.2004.05.007}.

\bibitem[Tu et~al.(2008)Tu, Narr, Doll{\'a}r, Dinov, Thompson, and
  Toga]{tu_brain_2008}
Z.~Tu, K.~L. Narr, P.~Doll{\'a}r, I.~Dinov, P.~M. Thompson, and A.~W. Toga.
\newblock Brain anatomical structure segmentation by hybrid
  discriminative/generative models.
\newblock \emph{{IEEE} Trans Med Imag}, 27\penalty0 (4):\penalty0 495--508,
  2008.
\newblock \doi{10.1109/TMI.2007.908121}.

\bibitem[Tustison et~al.(2010)Tustison, Avants, Cook, Zheng, Egan, Yushkevich,
  and Gee]{tustison_n4itk:_2010}
N.~J. Tustison, B.~B. Avants, P.~A. Cook, Y.~Zheng, A.~Egan, P.~A. Yushkevich,
  and J.~C. Gee.
\newblock {N4ITK:} improved {N3} bias correction.
\newblock \emph{{IEEE} Trans Med Imag}, 29\penalty0 (6):\penalty0 1310--1320,
  2010.
\newblock \doi{10.1109/TMI.2010.2046908}.

\bibitem[Van~Leemput et~al.(1999{\natexlab{a}})Van~Leemput, Maes, Vandermeulen,
  and Suetens]{van_leemput_automated_1999}
K.~Van~Leemput, F.~Maes, D.~Vandermeulen, and P.~Suetens.
\newblock Automated model-based bias field correction of {MR} images of the
  brain.
\newblock \emph{{IEEE} Trans Med Imag}, 18\penalty0 (10):\penalty0 885--896,
  1999{\natexlab{a}}.
\newblock \doi{10.1109/42.811268}.

\bibitem[Van~Leemput et~al.(1999{\natexlab{b}})Van~Leemput, Maes, Vandermeulen,
  and Suetens]{van_leemput_automated_1999-1}
K.~Van~Leemput, F.~Maes, D.~Vandermeulen, and P.~Suetens.
\newblock Automated model-based tissue classification of {MR} images of the
  brain.
\newblock \emph{{IEEE} Trans Med Imag}, 18\penalty0 (10):\penalty0 897--908,
  1999{\natexlab{b}}.
\newblock \doi{10.1109/42.811270}.

\bibitem[Van~Leemput et~al.(2001)Van~Leemput, Maes, Vandermeulen, Colchester,
  and Suetens]{van_leemput_automated_2001}
K.~Van~Leemput, F.~Maes, D.~Vandermeulen, A.~Colchester, and P.~Suetens.
\newblock Automated segmentation of multiple sclerosis lesions by model outlier
  detection.
\newblock \emph{{IEEE} Trans Med Imag}, 20\penalty0 (8):\penalty0 677--688,
  2001.
\newblock \doi{10.1109/42.938237}.

\bibitem[Van~Leemput et~al.(2003)Van~Leemput, Maes, Vandermeulen, and
  Suetens]{van_leemput_unifying_2003}
K.~Van~Leemput, F.~Maes, D.~Vandermeulen, and P.~Suetens.
\newblock A unifying framework for partial volume segmentation of brain {MR}
  images.
\newblock \emph{{IEEE} Trans Med Imag}, 22\penalty0 (1):\penalty0 105--119,
  2003.
\newblock \doi{10.1109/TMI.2002.806587}.

\bibitem[Vandewalle et~al.(2009)Vandewalle, Kovacevic, and
  Vetterli]{vandewalle_reproducible_2009}
P.~Vandewalle, J.~Kovacevic, and M.~Vetterli.
\newblock Reproducible research in signal processing.
\newblock \emph{{IEEE} Signal Process Mag}, 26\penalty0 (3):\penalty0 37--47,
  2009.
\newblock \doi{10.1109/MSP.2009.932122}.

\bibitem[Vovk et~al.(2007)Vovk, Pernus, and Likar]{vovk_review_2007}
U.~Vovk, F.~Pernus, and B.~Likar.
\newblock A review of methods for correction of intensity inhomogeneity in
  {MRI}.
\newblock \emph{{IEEE} Trans Med Imag}, 26\penalty0 (3):\penalty0 405--421,
  2007.
\newblock \doi{10.1109/TMI.2006.891486}.

\bibitem[Vrooman et~al.(2007)Vrooman, Cocosco, van~der Lijn, Stokking, Ikram,
  Vernooij, Breteler, and Niessen]{vrooman_multi-spectral_2007}
H.~A. Vrooman, C.~A. Cocosco, F.~van~der Lijn, R.~Stokking, M.~A. Ikram, M.~W.
  Vernooij, M.~M.~B. Breteler, and W.~J. Niessen.
\newblock Multi-spectral brain tissue segmentation using automatically trained
  k-nearest-neighbor classification.
\newblock \emph{{NeuroImage}}, 37\penalty0 (1):\penalty0 71--81, 2007.
\newblock \doi{10.1016/j.neuroimage.2007.05.018}.

\bibitem[West et~al.(2012)West, Warntjes, and Lundberg]{west_novel_2012}
J.~West, J.~Warntjes, and P.~Lundberg.
\newblock Novel whole brain segmentation and volume estimation using
  quantitative {MRI}.
\newblock \emph{Eur Radiol}, 22\penalty0 (5):\penalty0 998--1007, 2012.
\newblock \doi{10.1007/s00330-011-2336-7}.

\bibitem[Wright et~al.(1995)Wright, {McGuire}, Poline, Travere, Murray, Frith,
  Frackowiak, and Friston]{wright_voxel-based_1995}
I.~Wright, P.~{McGuire}, J.-B. Poline, J.~Travere, R.~Murray, C.~Frith,
  R.~Frackowiak, and K.~Friston.
\newblock A voxel-based method for the statistical analysis of gray and white
  matter density applied to schizophrenia.
\newblock \emph{{NeuroImage}}, 2\penalty0 (4):\penalty0 244--252, 1995.
\newblock \doi{10.1006/nimg.1995.1032}.

\bibitem[Yoo and Metaxas(2005)]{yoo_open_2005}
T.~S. Yoo and D.~N. Metaxas.
\newblock Open science - combining open data and open source software: Medical
  image analysis with the insight toolkit.
\newblock \emph{Med Image Anal}, 9\penalty0 (6):\penalty0 503--506, 2005.
\newblock \doi{10.1016/j.media.2005.04.008}.

\bibitem[Yushkevich et~al.(2006)Yushkevich, Piven, Hazlett, Smith, Ho, Gee, and
  Gerig]{yushkevich_user-guided_2006}
P.~A. Yushkevich, J.~Piven, H.~C. Hazlett, R.~G. Smith, S.~Ho, J.~C. Gee, and
  G.~Gerig.
\newblock User-guided {3D} active contour segmentation of anatomical
  structures: Significantly improved efficiency and reliability.
\newblock \emph{{NeuroImage}}, 31\penalty0 (3):\penalty0 1116--1128, 2006.
\newblock \doi{10.1016/j.neuroimage.2006.01.015}.

\bibitem[Zhang et~al.(2001)Zhang, Brady, and Smith]{zhang_segmentation_2001}
Y.~Zhang, M.~Brady, and S.~Smith.
\newblock Segmentation of brain {MR} images through a hidden markov random
  field model and the expectation-maximization algorithm.
\newblock \emph{{IEEE} Trans Med Imag}, 20\penalty0 (1):\penalty0 45--57, 2001.
\newblock \doi{10.1109/42.906424}.

\bibitem[Zijdenbos et~al.(1998)Zijdenbos, Evans, Riahi, Sled, Chui, and
  Kollokian]{zijdenbos_automatic_1998}
A.~Zijdenbos, A.~Evans, F.~Riahi, J.~Sled, J.~Chui, and V.~Kollokian.
\newblock Automatic quantification of multiple sclerosis lesion volume using
  stereotaxic space.
\newblock In \emph{Medical Image Computing and Computer-Assisted Intervention
  ({MICCAI)}}, vol. 1131 of \emph{{LNCS}}, pp. 439--448, Cambridge,
  Massachusetts, {US}, 1998. Springer Berlin Heidelberg.
\newblock \doi{10.1007/BFb0046984}.

\bibitem[Zijdenbos et~al.(2002)Zijdenbos, Forghani, and
  Evans]{zijdenbos_automatic_2002}
A.~Zijdenbos, R.~Forghani, and A.~Evans.
\newblock Automatic "pipeline" analysis of {3D} {MRI} data for clinical trials:
  application to multiple sclerosis.
\newblock \emph{{IEEE} Trans Med Imag}, 21\penalty0 (10):\penalty0 1280--1291,
  2002.
\newblock \doi{10.1109/TMI.2002.806283}.

\end{thebibliography}

\newpage
\onecolumn

\makeatletter
\ifbool{submit}
{}
{
  \efloat@restorefloats
}
\makeatother

\appendix
\setcounter{table}{0}
\setcounter{figure}{0}
\setcounter{algorithm}{0}
\renewcommand\thealgorithm{\Alph{section}.\arabic{algorithm}}
\makeatletter{}\begin{landscape}
\section{Data}\label{a:data}

\begin{table*}[!h]
\caption[Datasets]{Data resources used in this work\label{table:data}}
\rowcolors{2}{white}{lightgray}
\begin{minipage}{\linewidth}
\footnotesize
\setlength\LTleft{0pt}
\setlength\LTright{0pt}
\renewcommand\multirowsetup{\raggedright}
\makeatletter{}\begin{tabularx}{\linewidth}{>{\cellcolor{white}}l|>{\cellcolor{white}}l|>{\cellcolor{white}}l|L}
\multicolumn{1}{c}{Database}& \multicolumn{1}{c}{Cohort} & \multicolumn{1}{c}{Seq.} & \multicolumn{1}{c}{MR Parameters} \\
\hline
 & & T1w & Spoiled FLASH sequence, TR/TE=22/9.2ms, FA (flip angle)=30\degree, $181\times217\times181$ matrix size, 1mm isotropic voxel size. \\
 & & T2w & Dual echo spin echo (DSE), late echo sequence, TR=3300ms, TEs=35,120ms, FA=90$\degree$, $181\times217\times181$ matrix size, 1mm isotropic voxel size. \\
\multirow{-3}{*}{Brainweb$^{1}$} & \multirow{-3}{*}{1 dataset}  & PDw & Dual echo spin echo (DSE), early echo sequence, TR=3300ms, TEs=35,120ms, FA=90$\degree$, $181\times217\times181$ matrix size, 1mm isotropic voxel size. \\
\hline
 & & T1w & MPRAGE TR/TE/TI=6.7/3.1/842ms, FA=8\degree , $170\times256\times256$ matrix size, $1.0\times1.0\times1.2\,mm^3$ voxel size. \\
 & & T2w & Multi-shot turbo spin echo (TSE), TR/TE=2500/287ms, TSE/SENSE factors=100/2, fat suppression with SPIR, $180\times256\times256$ matrix size, $1.0\times0.9375\times0.9375\,mm^3$ voxel size. \\
 \multirow{-3}{*}{Kirby 21$^{2}$} & \multirow{-3}{*}{\parbox{3.4cm}{21 healthy volunteers \\ (11M/10F, 22-61yr.) }}
 & MT  & Spoiled 3D gradient echo, TR/TE=64/15ms, FA=9\degree, $256\times256\times95$ matrix size, $1.5\times1.5\times1.5\,mm^3$ voxel size.
 \\
\hline
& & & Philips Medical Systems Intera 3T, MPRAGE TR/TE = 9.6/4.6 ms, FA=8\degree. \\
& & & Philips Medical Systems Gyroscan Intera 1.5T, MPRAGE TR/TE = 9.81/4.6 ms, FA=8\degree. \\
& & \multirow{-3}{*}{T1w} & GE 1.5T System, information undisclosed by the time of accessing. \\
& & & Philips Medical Systems Intera 3T, TR/TE = 5725/100.0ms, FA=90\degree.\\
& & & Philips Medical Systems Gyroscan Intera 1.5T, 5725/100.0ms, FA=90\degree. \\
& & \multirow{-3}{*}{T2w} & GE 1.5T System, information undisclosed by the time of accessing. \\
& & & Philips Medical Systems Intera 3T, TR/TE = 5725/8.0ms, FA=90\degree. \\
& & & Philips Medical Systems Gyroscan Intera 1.5T, TR/TE = 8178/8.0ms, FA=90\degree. \\
& & \multirow{-3}{*}{PDw} & Philips Medical Systems Gyroscan Intera 1.5T, 5725/100.0ms, FA=90\degree. \\
\multirow{-9}{*}{IXI$^{3}$}
& \multirow{-9}{*}{465 healthy subjects.}
 & & GE 1.5T System, information undisclosed by the time of accessing. \\
\end{tabularx} 
$^{1}$ \url{http://www.bic.mni.mcgill.ca/brainweb/}, simulated dataset. \\
$^{2}$ \url{http://www.nitrc.org/projects/multimodal}, Achieva 3T Scanner (Philips Healthcare, Best, The Netherlands). \\
$^{3}$ \url{http://brain-development.org/index.php?n=Main.Datasets}, Multi-site study (several scanning systems). \\
\end{minipage}
\end{table*}
\end{landscape}

\newpage
\section{Optimization algorithms}\label{a:algorithms}

\begin{algorithm}[h]
\begin{enumerate}
\item[] \textbf{Initialization}.
Choose the best initialization for the model,
  $\lbrace \pi_{k,i}, \theta_k \rbrace^{(t=0)}$.
 
\bitem{Expectation Step}
Compute the \emph{posterior densities} $\gamma_{k,i}^{(t)}$ \eqref{eq:post_density}.
 
\bitem{Bias correction} Estimate $E$ as in \eqref{eq:bias_error} and perform the approximation
  of the smooth function, obtaining $B$.
Finally, set $\vy_i^{(t)} = \vy_i^{(t-1)} - \vec{b}_i^{(t)}$, to correct data for
  the last estimation of bias.
 
\bitem{Maximization Step}
  Estimate new parameters for the model, using the following update equations:
  
  \begin{equation*}
  \begin{aligned}
  \label{eq:m_step}
  \boldsymbol{\mu}_{k}^{(t)}&=\frac{1}{N_k} \, \underset{\forall i}\sum \gamma_{k,i}^{(t-1)} \, \vy_{i}\,, \\
  \boldsymbol{\Sigma}_{k}^{(t)}&=\frac{1}{N_k} \, \underset{\forall i}\sum \gamma_{k,i}^{(t-1)} \, (\vy_{i}-\boldsymbol{\mu}_{k}^{(t)})\,(\vy_{i}-\boldsymbol{\mu}_{k}^{(t)})^{T} \\
  \textrm{with} &\; N_k = \sum_{\forall i \in S} \gamma_{k,i}^{(t-1)}.
  \end{aligned}
  \end{equation*}

\item[] \textbf{Repeat} steps 1-3 until convergence.
\end{enumerate}
\caption{The \gls*{e_m} algorithm}\label{alg:e_m}
\end{algorithm}

\begin{algorithm}
\caption{$\alpha\beta$-Swap algorithm, as proposed by \cite{boykov_fast_2001}}
\label{alg:swap}
\begin{minipage}{\textwidth}
\begin{algorithmic}[1]
	\REQUIRE Arbitrary initial labeling $X$
	\STATE success $\leftarrow$ \FALSE
	\WHILE { success $\neq$ \TRUE }
		\FORALL{ pair of labels $\left\lbrace\alpha,\beta\right\rbrace$ } 
			\STATE Find $\hat{X}= \textrm{argmin} \left\lbrace E(X') \right\rbrace $ among $X'$ within one $\alpha-\beta$ swap of $X$
			\IF { $E(\hat{X}) < E({X})$ }
				\STATE $X \leftarrow \hat{X}$
			\ELSE
				\STATE success $\leftarrow$ \TRUE
			\ENDIF
		\ENDFOR
	\ENDWHILE
	\RETURN $f$
\end{algorithmic}
\noindent \rule{\textwidth}{0.4pt}

\footnotesize
\noindent
Given a label $\alpha$, an $\alpha$-expansion move is a change of a number
  of image pixels from any original label to $\alpha$.
Equivalently, given a pair of labels $\alpha,\beta$, an
  $\alpha\beta$-swap is a move where a number of pixels 
  with label $\alpha$ change to $\beta$ and 
  a number of pixels previously labeled $\beta$ change to $\alpha$.
\end{minipage}
\end{algorithm} 

\ifbool{submit}
{}
{
\newpage
\setcounter{figure}{0} \renewcommand\thefigure{\arabic{figure}}
\setcounter{table}{0} \renewcommand\thetable{\arabic{table}}
}

\end{document}